\documentclass[nohyperref]{article}

\usepackage{microtype}
\usepackage{graphicx}
\usepackage{booktabs} 
\usepackage{subcaption}
\usepackage{multirow}

\usepackage{hyperref}
\usepackage[table]{xcolor}

\usepackage[accepted]{icml2023}

\usepackage{amsmath}
\usepackage{amssymb}
\usepackage{mathtools}
\usepackage{amsthm}

\usepackage[capitalize,noabbrev]{cleveref}

\theoremstyle{plain}
\newtheorem{theorem}{Theorem}[section]

\theoremstyle{definition}
\newtheorem{definition}[theorem]{Definition}

\theoremstyle{remark}

\usepackage[textsize=tiny]{todonotes}

\icmltitlerunning{SEGA: Structural Entropy Guided Anchor View for Graph Contrastive Learning}

\begin{document}

\twocolumn[
\icmltitle{SEGA: Structural Entropy Guided Anchor View \\
           for Graph Contrastive Learning}



\icmlsetsymbol{equal}{*}

\begin{icmlauthorlist}
\icmlauthor{Junran Wu}{nlsde,equal}
\icmlauthor{Xueyuan Chen}{nlsde,equal}
\icmlauthor{Bowen Shi}{nlsde}
\icmlauthor{Shangzhe Li}{nlsde}
\icmlauthor{Ke Xu}{nlsde,zhongguancun}
\end{icmlauthorlist}

\icmlaffiliation{nlsde}{State Key Lab of Software Development Environment, Beihang University, Beijing, 100191, China}

\icmlaffiliation{zhongguancun}{Zhongguancun Laboratory, Beijing 100094, China}

\icmlcorrespondingauthor{Shangzhe Li}{shangzheli@buaa.edu.cn}

\icmlkeywords{Machine Learning, ICML}

\vskip 0.3in
]



\printAffiliationsAndNotice{\icmlEqualContribution} 

\begin{abstract}
In contrastive learning, the choice of ``view'' controls the information that the representation captures and influences the performance of the model. 
However, leading graph contrastive learning methods generally produce views via random corruption or learning, which could lead to the loss of essential information and alteration of semantic information.
An anchor view that maintains the essential information of input graphs for contrastive learning has been hardly investigated.
In this paper, based on the theory of graph information bottleneck, we deduce the definition of this anchor view; put differently, \textit{the anchor view with essential information of input graph is supposed to have the minimal structural uncertainty}.
Furthermore, guided by structural entropy, we implement the anchor view, termed \textbf{SEGA}, for graph contrastive learning.
We extensively validate the proposed anchor view on various benchmarks regarding graph classification under unsupervised, semi-supervised, and transfer learning and achieve significant performance boosts compared to the state-of-the-art methods.
\end{abstract}

\section{Introduction}
Self-supervised learning has gained popularity recently and achieved great success in deep learning, $e.g.,$ BERT~\cite{devlin2019bert} and MoCo~\cite{he2020momentum}. Compared with supervised learning, self-supervised learning gets equal or even better performance with limited or no-labeled data which saves much annotation time and plenty of resources. As one of the empirical self-supervised learning methods, contrastive learning develops rapidly and recently has been applied to various domains because of the scarcity of datasets with labels.
Meanwhile, graph neural networks (GNNs) have become ubiquitous for graphs because of their ability to model structural information~\cite{li2022chart,zhang2022hierarchical}.
Therefore, graph contrastive learning~\cite{you2020graph,suresh2021adversarial,yang2022omni}, based on the success of contrastive learning in computer vision and natural language processing, has attracted plenty of research interest after its presentation.

In contrastive learning, the choice of view controls the information that the representation captures. Researchers have found that the quality of views influences the performance of contrastive learning models~\cite{tian2020makes} and focus on the generation of effective views that lead to better performance for graphs through the data augmentation~\cite{suresh2021adversarial,you2021graph}. 
Despite the effectiveness of these graph views on various tasks, the proposed data augmentations via random corruption or learning suffer from structural damage and artificially introduced noise, which could alter the fundamental property of input graphs.
Unlike images, data augmentation on graphs is much harder to provide high-quality contrastive samples due to the rich structural information of various contexts in the graph data~\cite{feng2022adversarial}.
So far, little attention has been paid to the anchor view for graph contrastive learning that maintains the essential information of input graphs regarding graph classification.
Therefore, we are eager to provide high-quality contrastive samples by settling the two questions: \textit{(1) What is the anchor view holding essential information? (2) How to generate the anchor view for graph contrastive learning?}

\begin{figure*}[ht]
\vskip 0.2in
\begin{center}
\centerline{\includegraphics[width=\textwidth]{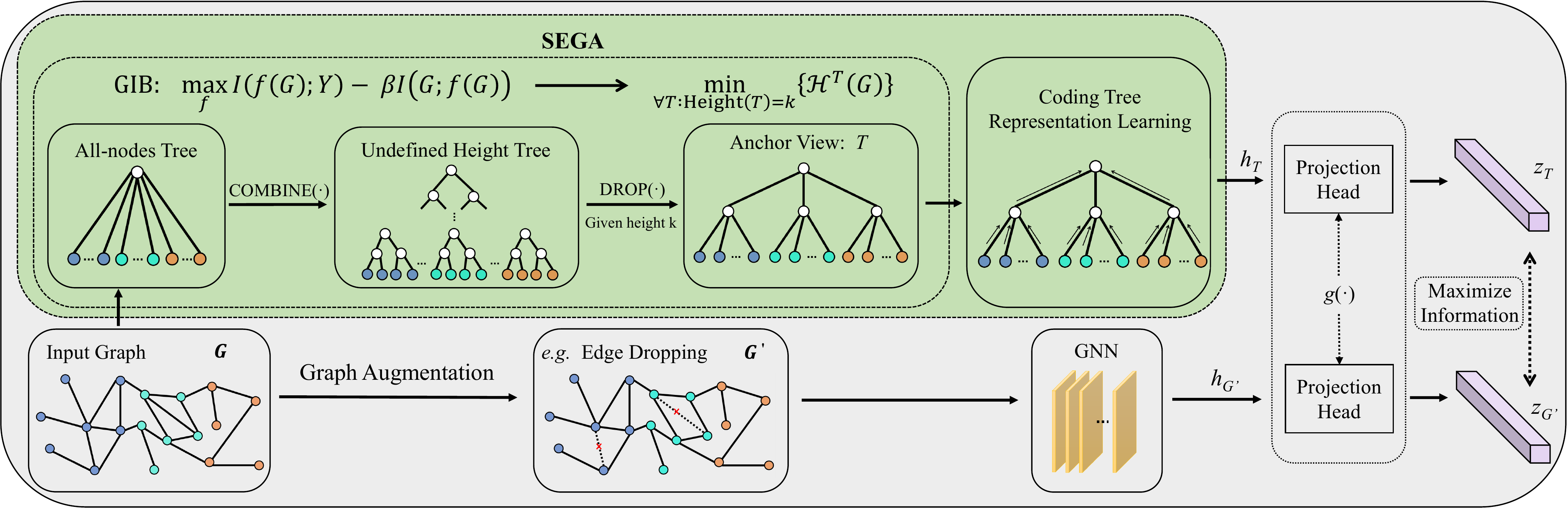}}
\caption{\textbf{Graph contrastive learning framework with SEGA.} The anchor view with essential information for graph contrastive learning is realized with the graph information bottleneck and structural entropy. The essential information of given graph is decoded to corresponding coding tree by structural entropy minimization. 
Under the architecture of contrastive learning, the model is trained to maximize the consensus between the coding tree representation $h_T$ and the augmentation-based graph embedding $h_{G'}$.}
\label{fig:framework}
\end{center}
\vskip -0.2in
\end{figure*}

Recently, the information bottleneck theory that encourages model to capture minimal but sufficient information, that is essential information, has been applied to learn graph representation, called Graph Information Bottleneck (GIB)~\cite{wu2020graph}.
In light of GIB, we conclude that \textit{the anchor view with essential information of input graph is supposed to have the minimal structural uncertainty} ($i.e., G^\ast=\min\mathcal{H}(G)$).
Now, with the definition of target anchor view, the last question is its instantiation for graph contrastive learning.
Thus, a metric for graph structural uncertainty measurement is needed.
Recently, based on the classic uncertainty metric, Shannon entropy~\cite{shannon1948mathematical}, researchers proposed the structural entropy to measure the uncertainty of graph structures~\cite{li2016structural}.
This theory implies that under intuition, people fear uncertainty and usually follow the choices that minimize such non-determinism.
According to the structural information theory, the essential information embedded in the corresponding graph can be decoded by minimizing its structural uncertainty, that is to say, minimizing the structural entropy.

Here, in view of the definition of the anchor view given above, we propose its instantiation, termed \textbf{SEGA} (see~\cref{fig:framework}), guided by structural entropy minimization for graph contrastive learning.
Specifically, an optimization algorithm is first introduced for structural entropy minimization, in which the coding trees of corresponding graphs are generated for essential information extraction. Then, based on the message-passing scheme in GNNs, an encoder is proposed to obtain the essential information held by the transformed coding trees. Contrasted with the effective views in previous works, extensive experiments, including unsupervised, semi-supervised, and transfer learning, are conducted on various benchmarks regarding graph classification. Superior performance can be observed in comparison with those state-of-the-art (SOTA) methods. The contributions of this work can be summarized as follows:

\begin{itemize}
\item Based on the theory of graph information bottleneck, to the best of our knowledge, we are the first to figure out the anchor view with essential information of input graphs for graph contrastive learning. 

\item Guided by structural entropy minimization, we present an instantiation, termed SEGA, to implement the proposed anchor view for graph contrastive learning.

\item We extensively evaluate the proposed anchor view on various benchmarks under the setting of unsupervised, semi-supervised, and transfer learning, and obtain superior performance compared to the SOTA methods.
\end{itemize}

\section{Related Works}
\textbf{Graph Contrastive Learning.} Great success has been achieved by graph contrastive learning when facing the label scarcity in real-world network data \cite{you2020graph,suresh2021adversarial, you2021graph,feng2022adversarial,you2022bringing}.
However, unlike the data augmentation on images that do not require rich domain knowledge, the graph augmentation is far less intuitive and hard to analyze, which makes it difficult to produce high-quality contrast samples \cite{feng2022adversarial,you2022bringing}.
Hence, the study of the graph contrastive view is a key issue in graph contrastive learning.
Recently, based on the data augmentation on images, plenty of efforts have been devoted to exploring various augmentations on graphs \cite{you2020graph,you2021graph,suresh2021adversarial,you2022bringing,li2022let}. 
While effective, none of them try to identify the essential information from graphs.
Furthermore, the proposed data augmentations via random perturbation or learning suffer from structure damage and noisy information \cite{you2020graph, suresh2021adversarial}.
Recently, besides the view exploration, GraphLoG~\cite{xu2021self} and OEPG~\cite{yang2022omni} are built upon the generic graph contrastive learning methods to discover the global semantic structure underlying the whole dataset. Although they present excellent performance, in this work, we are devoted to the domain of view generation, which is orthogonal to the works for dataset semantic structure exploration; put differently, extensive works for contrastive view generation can work with the framework of GraphLoG and OEPG to produce more superior performance.

\textbf{Structural Entropy.} 
Information entropy, as the basis of Information Theory, stems from the demand for information measuring in communication systems \cite{shannon1948mathematical}. Considering measuring the information in graphs, lots of metrics were proposed. The entropy of graphs can be measured with $p(G)$ at the global level~\cite{mowshowitz2012entropy}. For a signal graph, various works aim to measure the structural entropy of nodes. A local measurement of graph entropy was first proposed based on distance \cite{raychaudhury1984discrimination}. Subsequently, extensive researches attempted to measure the structural information of graph from different angles, such as Von Neumann entropy \cite{braunstein2006laplacian}, parametric graph entropy \cite{dehmer2008information}, Gibbs entropy \cite{bianconi2009entropy}. 
However, these definitions all destructure the graph into an unstructured probability distribution and then apply Shannon entropy to define the information of the graph. Thus, these metrics can not serve as the measurement of structural information that is crucial for graphs and the key to the success of GNNs.
Recently, based on coding trees, structural entropy was proposed to evaluate the complexity of the hierarchical structure of a graph \cite{li2016structural}. Considering the measurement of graph information with fixed hierarchical manners, $k$-dimensional structural entropy was further defined and can be used to decode the essential information of graphs \cite{li2018decoding,wu2022structural,wu2022simple}.

\section{Notations and Preliminaries}
Some preliminary concepts and notations are introduced here. In this work, $\mathbb{G} = \{G_1,G_2,\cdots,G_M\}$ refers to a set of graphs, and each graph can be represented as a two tuple $G=(\mathcal{V}, \mathcal{E})$, where $\mathcal{V}$ and $\mathcal{E}$ are the sets of nodes and edges. 

\textbf{Graph representation learning.} In this work, GNNs with message-passing scheme are adopted as the encoders. A GNN aims to learn an embedding vector $h_v\in\mathbb{R}$ for each node and a vector $h_G\in\mathbb{R}$ for the entire graph $G$. A node representation $h_v$ is initialized as $h_v^{(0)}=X_v$, and will be iteratively updated by an encoder. For an $L$-layer GNN, each node presentation will be updated using $L-$hop information from surrounding nodes. The $l$-layer of a GNN~\cite{gilmer2017neural} can be expressed as
\begin{equation}
h_v^{(l)} = f_U^{(l)}(h_v^{(l\text{-}1)},f_M^{(l)}(\{(h_v^{(l\text{-}1)},h_u^{(l\text{-}1)})|u\in N(v)\})),
\end{equation}
where $N(v)$ is the neighborhood node set for $v$, $h_v^{(l)}$ is the node representation of $v$ in the $l$-th layer, $f_U^{(l)}$ denotes the update function in the $l$-th layer, and $f_M^{(l)}$ refers to the trainable message-passing function in the $l$-th layer. $h_v$ can be referred to a summary of neighborhood nodes, like a subgraph. Thus, after $L$ iterations, the entire graph representation can be formalized as follows:
\begin{equation}
h_G = f_R(\{h_v|v\in\mathcal{V}\}),
\end{equation}
where $f_R$ is the readout function which pools the final set of node representations.

\textbf{Graph contrastive learning.}
In a generic contrastive learning model for graph classification, two corresponding views of the same graph $G_i$ are generally generated by two data augmentation operators and serve as a positive pair.
Let $\tilde{G}^1_i$ and $\tilde{G}^2_i$ be the two augmented views; then, a GNN-based encoder is adopted to model the structural information underlying the given graph. In the pre-training phase, a projection head is further employed to map the two views into an embedding space for contrasting.
The released feature vectors $h_i^1$ and $h_i^2$ are designed to identify themselves from the others.
Correspondingly, the NT-Xent loss \cite{chen2020simple} helps to achieve the goal of graph contrastive learning that maximizes the consensus of two correlated views:
\begin{equation}
\mathcal{L}_i = -\log\frac{e^{sim(h_i^1, h_i^2)/\tau}}{\sum^N_{j=1,j\neq i}e^{sim(h_i^1, h_j^2)/\tau}},
\end{equation}
where $N$ denotes the batch size, $\tau$ is the temperature parameter, and $sim(h^1, h^2)$ is generally implemented by a cosine similarity function $\frac{h^{1\top} h^2}{||h^1||\cdot||h^2||}$.

\section{Methodology}
In this section, we first introduce our theoretical motivation and try to give the definition of the anchor view with essential information. Based on the structural information theory, we then present an instantiation of the anchor view for graph contrastive learning.

\subsection{The Anchor View Holding Essential Information}
Based on the idea of information bottleneck, GIB~\cite{wu2020graph} presents a statement that motivates us to think about the anchor view that maintains the essential information of input graphs. Specifically, through maximizing the mutual information (MI) between the output and target (i.e., $\max I(f(G); Y)$) while stinting such information between the input and output (i.e., $\min I(G; f(G)$), models are capable of learning minimal but sufficient information for a given task.
Therefore, the formal description of essential information is given by
\begin{definition}
The \textbf{essential information} learned from the input graphs is supposed to be the minimal sufficient information required for a downstream prediction task.
\end{definition}
In computer vision, researchers empirically gave a similar answer for contrastive learning; put differently, compressing the mutual information between views while maintaining the integrity of information related to downstream tasks \cite{tian2020makes}, which further convinces us to build the target anchor view via the GIB.
Formally, the objective of graph information bottleneck can be written as
\begin{equation}
\text{GIB:}\,\,\,\, \mathop{\max} I(f(G); Y) - \beta I(G; f(G)),
\label{eq:gib}
\end{equation}
where $(G, Y) \sim \mathbb{P}_{\mathcal{G} \times \mathcal{Y}}$ and $\beta>0$. 

Here, we give the definition of the target view:
\begin{definition}
The anchor view for graph contrastive learning is supposed to have minimal but sufficient information for downstream tasks and help the other view learn such information during training.
\end{definition}

In the light of GIB, to acquire the essential information, the anchor view for graph contrastive learning should have minimal but sufficient information for downstream tasks.
However, unfortunately, the first part of GIB requires the target-relevant information of the given task (i.e., $Y$), and as we know, it is impossible under the architecture of self-supervised training.~\footnote{Based on the property of structural entropy, the information for downstream tasks within the proposed anchor view $G^\ast$ is still more than views from augmentations. Details refer to Theorem~\ref{theo:anchor-lowbound}.}
In this context, the other part that does not require such target-related information sheds light on the path of the target anchor view exploration.
Therefore, minimizing the mutual information between the learned representation and input graph (i.e., $\min I(G; f(G))$) suggests the essential information that graph contrastive learning should conquer. 
Formally, we have
\begin{align}
\text{GIB:} &\,\,\mathop{\max} I(f(G); Y) - \beta I(G; f(G)), \nonumber \\ 
\Rightarrow &\,\,\min I(G; f(G)).
\end{align}
Here, we first give a property that the target anchor view should own:
\begin{definition}
The anchor view with essential information is supposed to be a substructure of the given graph to avoid artificially introduced noise.
\label{def:view_property}
\end{definition}

In computer vision, to obtain the essential information, data augmentation via random perturbation has been ubiquitously adopted, and the accompanying induced noise is also approved for robust representation learning~\cite{tian2020makes}.
However, unlike the data augmentation on images that does not require rich domain knowledge, the graph augmentation is far less intuitive and hard to analyze, which makes it difficult to produce high-quality contrast samples~\cite{feng2022adversarial,you2022bringing}.
Thus, we argue that the anchor view with essential information of given graphs should avoid the artificially introduced noise from random perturbation.
Now, let $G^\ast$ be the target anchor view  of a graph $G$, the mutual information between $G$ and $G^\ast$ can be formulated as
\begin{equation}
I(G^\ast; G) = \mathcal{H}(G^\ast) - \mathcal{H}(G^\ast|G),
\end{equation}
where $\mathcal{H}(G^\ast)$ is the entropy of $G^\ast$ and $\mathcal{H}(G^\ast|G)$ is the conditional entropy of $G^\ast$ conditioned on $G$.~\footnote{We omit the graph encoder $f$ for simplicity.}

\begin{theorem}
According to~\cref{def:view_property}, the information in $G^\ast$ is a subset of information in $G$ (i.e., $\mathcal{H}(G^\ast) \subseteq \mathcal{H}(G)$); thus, we have:
\begin{equation}
\mathcal{H}(G^\ast|G)=0.
\end{equation}
\label{theo:hgeq0}
\vspace{-10pt}
\end{theorem}

The detailed proof of \cref{theo:hgeq0} is shown in Appendix A. Here, the mutual information between $G$ and $G^\ast$ can be rewritten as
\begin{equation}
I(G^\ast; G) = \mathcal{H}(G^\ast).
\end{equation}
Accordingly, to acquire the anchor view with essential information, we need to optimize:

\begin{equation}
\min\,I(G; f(G)) \Rightarrow \min\,\mathcal{H}(G^\ast).
\end{equation}

Besides information measuring, $\mathcal{H}(G^\ast)$ also reveal the uncertainty of $G^\ast$ based on the definition of Shannon entropy \cite{shannon1948mathematical}.
Correspondingly, the definition of the anchor view with essential information for graph contrastive learning is given by:
\begin{definition}
The anchor view with essential information of input graph is supposed to have the minimal structural uncertainty.
\end{definition}

\noindent \textbf{Remark.} In this paper, base on GIB theory~\cite{wu2020graph}, we conclude that the anchor view with essential information is supposed to satisfy $\mathcal{H}(G^\ast)$, and we interpret it from the topological perspective as the view with minimal structural uncertainty. First, this is because topological structures are ubiquitous in complex network systems, while node features may not be always available. For example, in the social network datasets adopted in this study, only topological structures are available and node features are absent. Furthermore, current research on contrastive views mainly focuses on data augmentation methods based on topological structures such as edge perturbation, node dropout and subgraph extraction in GraphCL~\cite{you2020graph}; learning edge dropout in AD-GCL~\cite{suresh2021adversarial}; and learning node selection in RGCL~\cite{li2022let}. Therefore, we hope to explore contrastive views with essential information from a more general angle, that is topological structure.

Next, we will elaborate on the instantiation of the defined anchor view by introducing structural information theory.

\subsection{Instantiation of Anchor View}
In this subsection, we are going to introduce a practical instantiation, a structural entropy guided anchor view (i.e., \textbf{SEGA}), for essential information decoding.

Despite the broader applicability of Shannon entropy, in this work, we need the metric of structural uncertainty for graphs, which has also been asked by Brooks in the ``Three great challenges for half-century-old computer science''~\cite{brooks2003three}.
The question is how to define the underlying information of a graph so that the essential information of the graph could be decrypted, while Shannon wondered the feasibility of communication graph analysis via a structural theory of information \cite{shannon1953lattice}.
Recently, structural entropy defined on graphs was proposed to measure the uncertainty of the graph structure~\cite{li2016structural}. According to this structural information theory, a graph is encoded by a coding tree.~\footnote{A detailed description and illustration for coding tree $T$ on given graph $G$ can be found in Appendix B.}
The structural entropy of graph $G=(\mathcal{V}, \mathcal{E})$ on a coding tree $T$ is defined as
\begin{equation}
\label{eq:structural_entropy}
\mathcal{H}^T (G)=-\sum_{v_\tau \in T} \frac{g_{v_\tau}}{vol(\mathcal{V})} \log \frac{vol(v_\tau)}{vol(v_\tau^+)},
\end{equation}
where $v_\tau$ is a nonroot node in $T$ and represents a node subset $\mathcal{V}_\tau \subset \mathcal{V}$ based on its covered leaf nodes, $g_{v_\tau}$ is the number of edges with exactly one vertex in $\mathcal{V}_\tau$, $v_\tau^+$ refers to the immediate predecessor of $v_\tau$, and $vol(\mathcal{V})$, $vol(v_\tau)$ and $vol(v_\tau^+)$ are the sums of degrees of vertices in $\mathcal{V}$, $v_\tau$ and $v_\tau^+$, respectively. 
Thus, to decode the essential information of graph $G$ with minimal structural uncertainty, we need to realize the optimal coding tree $T$ with minimum entropy (i.e., $\min_{T}\mathcal{H}^T (G)$). 
Besides the optimal coding tree, considering that a real-world network generally has a natural structure with a fixed hierarchy, a coding tree with the corresponding fixed height is preferred.
In this context, $k$-dimensional structural entropy is used to decode the optimal coding tree with a certain height $k$:
\begin{equation}
\mathcal{H}^{k}(G)=\min_{\forall T:\text{Height}(T)=k}\{\mathcal{H}^T (G)\}.
\end{equation}
Now, in light of the structural information theory, we know that the essential information of input graph can be decoded by minimizing its structural entropy. Moreover, the target anchor view for graph contrastive learning with minimum but sufficient information is the coding tree of the graph through $k$-dimensional structural entropy minimization.

\begin{theorem}
\label{theo:anchor-lowbound}
Given the target anchor view $G^\ast$ and a general data augmentation function $t$, we have
\begin{equation}
I(G^\ast; Y) \ge I(t(G); Y).
\end{equation}
\end{theorem}

\begin{proof}
Suppose the encoder $f$ is implemented by a GNN. The optimal encoder $f^\ast$ is the best model which GNN can find. According to the definition $f^\ast = argmax_{f} I(f(G);G)$, $f^\ast$ should be injective. Given the target anchor view $G^\ast$, $G^\ast \Rightarrow f^\ast(G^\ast)$ is an injective deterministic mapping. Thus, for any random variable $Q$, 
\begin{equation}
I(f^\ast(G^\ast); Q) = I(G^\ast; Q). 
\end{equation}
When there is $Q = Y$, we will have,
\begin{equation}
I(f^\ast(G^\ast); Y) = I(G^\ast; Y).
\end{equation}
In light of the property of structural information theory~\cite{li2016structural}, structural entropy decodes the essential structure of the original system while measuring the structural information to support the semantic analysis of the system. Thus, we have
\begin{equation}
I(f^\ast(G^\ast); Y ) = I(f^\ast(G); Y). 
\end{equation}
Now, introducing the data processing inequality \cite{thomas2006elements} for data augmentation,
\begin{align}
I(f^\ast(G); Y) &= I(G; Y) \nonumber \\
& \geq I(t(G); Y) = I(f^\ast(t(G)); Y). 
\end{align}
Combining above equations, we can have 
\begin{align}
I(G^\ast; Y) & = I(f^\ast (G^\ast); Y) \nonumber \\ 
& = I(f^\ast(G); Y) \nonumber \\ 
& \geq I(f^\ast(t(G)); Y ) = I(t(G); Y ),
\label{eq:low-bound}
\end{align}
which concludes the proof.
\end{proof}

\cref{theo:anchor-lowbound} guarantees a lower bound of the mutual information between the learned representations and the labels of the downstream task; put differently, the learned essential information with the anchor view $G^\ast$ is more than views from augmentations.

\begin{algorithm}[!b]
\caption{Structural uncertainty minimization}
\label{code:coding_tree} 
\textbf{Input:} the given height $k>1$, and the candidate graph $G=(\mathcal{V}, \mathcal{E})$\\
\textbf{Output:} a coding tree $T$ that meets the height bar

\begin{algorithmic}[1]
\STATE Build a coding tree $T$ with a root node $v_r$ and all nodes in $\mathcal{V}$ as its children;
\STATE // Stage 1: Construct a full-height binary coding tree from bottom to top;
\WHILE{$|v_r.children|>2$} {
  \STATE $\text{COMBINE}(v^1_c, v^2_c)\leftarrow argmax_{(v^1_c, v^2_c)}\{\mathcal{H}^T(G) - \mathcal{H}^{T_{\text{COMBINE}(v^1_c, v^2_c)}}(G)| v^1_c, v^2_c \in v_r.children\}$;
}
\ENDWHILE
\STATE // Stage 2: Squeeze $T$ to meet the height bar;
\WHILE{$\text{Height}(T)>k$} {
  \STATE $\text{DROP}(v_\tau) \; \leftarrow \; argmin_{v_\tau}\{\mathcal{H}^{T_{\text{DROP}(v_\tau)}}(G) - \mathcal{H}^T(G)|v_\tau\in T\,\&\ v_\tau\neq v_r \,\&\, v_\tau\notin \mathcal{V}\}$;
}
\ENDWHILE
\STATE return $T$;
\end{algorithmic} 
\end{algorithm}

For structural entropy minimization, we aim to decrypt the coding tree with fixed height $k$ from a graph. Given a graph $G=(\mathcal{V}, \mathcal{E})$, a coding tree $T$ can be build, in which $v_r$ is the root node of $T$ and $\mathcal{V}$ are the leaf nodes of $T$. For the coding tree $T$, there are two function definitions.
\begin{definition}
\label{def:merge}
Given any two child nodes of $v_r$, $v^1_c$ and $v^2_c$, there is a function $\text{COMBINE}(v^1_c, v^2_c)$ for $T$ to add a new node $v_i$ between $v_r$ and $(v^1_c, v^2_c)$:
\begin{align}
  v_i.children &= \{v^1_c,v^2_c\}, \\
  v_r.children &= \{v_i\}\cup v_r.children.
\end{align}
\end{definition}

\begin{definition}
Given a pair of nodes $(v_\tau, v^+_\tau)$ in $T$, there is a function $\text{DROP}(v_\tau)$ for $T$ to drop node $v_\tau$ and fuse the children of $v_\tau$ into $v^+_\tau$:
\begin{equation}
  v^+_\tau.children = v^+_\tau.children\cup v_\tau.children.
\end{equation}
\end{definition}

Based on the two defined operators, \cref{code:coding_tree} shows the realization of structural uncertainty minimization. 
Specifically, given a coding tree including only root node and leaf nodes, a full-height binary coding tree grows from bottom to top. In this process, two child nodes of root are combined to form a new division in each iteration, which aims to minimize the structural entropy. $T_{\text{COMBINE}(v_c^1, v_c^2)}$ denotes the coding tree that has combined the two children (i.e., $v_c^1$ and $v_c^2$) of the root node.
Then, considering the height limitation, the well developed coding tree needs to be squeezed. 
During each iteration, an inner-node from $T$ will be dropped until its height meets the bar. In particular, each dropped node should ensure that $T$ has the minimized structural entropy after each iteration. $T_{\text{DROP}(v_\tau)}$ is the coding tree that has dropped the inner node $v_\tau$. In the end, a fixed height coding tree $T=(\mathcal{V}^T, \mathcal{E}^T)$ will be obtained, in which $\mathcal{V}^T = (\mathcal{V}_0^T,\dots,\mathcal{V}_k^T)$ and $\mathcal{V}_0^T = \mathcal{V}$. 
The running process of~\cref{code:coding_tree} is illustrated in Appendix B.

\textbf{Anchor view representation learning.}
Having the algorithm for structural uncertainty minimization, we are capable of producing the anchor view with essential information for graph contrastive learning. 
To further integrate the coding tree into the architecture of contrastive learning, we give an encoder for the anchor view representation learning. 
In light of the graph convolution scheme in GNNs, the coding tree encoder is designed to iteratively transfer messages from bottom to top.
Specifically, based on the hierarchical structure of the coding tree and the initial node feature of leaves, the non-leaf nodes update their hidden representation by aggregating the hidden features from their children. 
Formally, the $i$-th layer of the encoder can be written as, $x_v^i = \text{MLP}^i\left(\sum\nolimits_{u\in\mathcal{L}(v)}x_u^{(i-1)}\right)$,
where $x^i_v$ is the feature of $v$ in the $i$-th layer of coding tree $T$, $x^0_v$ is the input feature of leaf nodes, and $\mathcal{L}(v)$ refers to the children of $v$.

\begin{table*}[!t]
\centering
\caption{Average accuracies (\%) $\pm$ Std. of compared methods via unsupervised learning. \textbf{Bold} indicates the best performance over all methods. A.A. refers to the average accuracy over eight benchmarks. A.R. implies the abbreviation of average rank. The results of baselines are derived from the published works and - indicates the data missing in the such works.}
\label{tab:unsupervised}
\vskip 0.15in
\begin{small}
\resizebox{\textwidth}{!}{%
\begin{tabular}{lcccc|cccc|cc}
\hline \hline
 & NCI1 & PROTEINS & DD & MUTAG & COLLAB & RED-B & RED-M5K & IMDB-B & A.A. & A.R. \\ \hline \hline
GL & - & - & - & 81.66$\pm$2.11 & - & 77.34$\pm$0.18 & 41.01$\pm$0.17 & 65.87$\pm$0.98 & - & 6.5 \\
WL & 80.01$\pm$0.50 & 72.92$\pm$0.56 & - & 80.72$\pm$3.00 & - & 68.82$\pm$0.41 & 46.06$\pm$0.21 & 72.30$\pm$3.44 & - & 5.3 \\
DGK & \textbf{80.31$\pm$0.46} & 73.30$\pm$0.82 & - & 87.44$\pm$2.72 & - & 78.04$\pm$0.39 & 41.27$\pm$0.18 & 66.96$\pm$0.56 & - & 4.3 \\ \hline
node2vec & 54.89$\pm$1.61 & 57.49$\pm$3.57 & - & 72.63$\pm$10.20 & - & - & - & - & - & 7.7 \\
sub2vec & 52.84$\pm$1.47 & 53.03$\pm$5.55 & - & 61.05$\pm$15.80 & - & 71.48$\pm$0.41 & 36.69$\pm$0.42 & 55.26$\pm$1.54 & - & 8.5 \\
graph2vec & 73.22$\pm$1.81 & 73.30$\pm$2.05 & - & 83.15$\pm$9.25 & - & 75.78$\pm$1.03 & 47.86$\pm$0.26 & 71.10$\pm$0.54 & - & 5.3 \\
MVGRL & - & - & - & 75.40$\pm$7.80 & - & 82.00$\pm$1.10 & - & 63.60$\pm$4.20 & - & 6.7 \\
InfoGraph & 76.20$\pm$1.06 & 74.44$\pm$0.31 & 72.85$\pm$1.78 & 89.01$\pm$1.13 & 70.65$\pm$1.13 & 82.50$\pm$1.42 & 53.46$\pm$1.03 & 73.03$\pm$0.87 & 74.02 & 2.9 \\
GraphCL & 77.87$\pm$0.41 & 74.39$\pm$0.45 & 78.62$\pm$0.40 & 86.80$\pm$1.34 & 71.36$\pm$1.15 & 89.53$\pm$0.84 & 55.99$\pm$0.28 & 71.14$\pm$0.44 & 75.71 & 2.9 \\ \hline
SEGA & 79.00$\pm$0.72 & \textbf{76.01$\pm$0.42} & \textbf{78.76$\pm$0.57} & \textbf{90.21$\pm$0.66} & \textbf{74.12$\pm$0.47} & \textbf{90.21$\pm$0.65} & \textbf{56.13$\pm$0.30} & \textbf{73.58$\pm$0.44} & \textbf{77.25} & \textbf{1.3} \\ \hline \hline
\end{tabular}%
}
\end{small}
\vskip -0.1in
\end{table*}

\begin{table*}[!t]
\centering
\caption{Average test ROC-AUC (\%) $\pm$ Std. over different 10 runs of SEGA along with all baselines on nine downstream benchmarks. The results of baselines are derived from the corresponding works. \textbf{Bold} indicates the best performance among all baselines. Avg. shows the average ROC-AUC over all datasets.}
\label{tab:trans-results}
\vskip 0.15in
\begin{small}
\resizebox{\textwidth}{!}{%
\begin{tabular}{l|cccccccc|c|c}
\hline \hline
 & BBBP & Tox21 & ToxCast & SIDER & ClinTox & MUV & HIV & BACE &  PPI & Avg. \\ \hline \hline
No Pre-Train & 65.8$\pm$4.5   & 74.0$\pm$0.8   & 63.4$\pm$0.6   & 57.3$\pm$1.6   & 58.0$\pm$4.4   & 71.8$\pm$2.5   & 75.3$\pm$1.9   & 70.1$\pm$5.4   & 64.8$\pm$1.0 & 66.72 \\ \hline
Infomax      & 68.8$\pm$0.8   & 75.3$\pm$0.6   & 62.7$\pm$0.4   & 58.4$\pm$0.8   & 69.9$\pm$3.0   & 75.3$\pm$2.5   & 76.0$\pm$0.7   & 75.9$\pm$1.6   & 64.1$\pm$1.5 & 69.60 \\
EdgePred     & 67.3$\pm$2.4   & 76.0$\pm$0.6   & 64.1$\pm$0.6   & 60.4$\pm$0.7   & 64.1$\pm$3.7   & 74.1$\pm$2.1   & 76.3$\pm$1.0   & \textbf{79.9$\pm$0.9}   & 65.7$\pm$1.3 & 69.76 \\
AttrMasking  & 64.3$\pm$2.8   & 76.7$\pm$0.4   & 64.2$\pm$0.5   & 61.0$\pm$0.7   & 71.8$\pm$4.1   & 74.7$\pm$1.4   & 77.2$\pm$1.1   & 79.3$\pm$1.6   & 65.2$\pm$1.6 & 69.38 \\
ContextPred  & 68.0$\pm$2.0   & 75.7$\pm$0.7   & 63.9$\pm$0.6   & 60.9$\pm$0.6   & 65.9$\pm$3.8   & 75.8$\pm$1.7   & 77.3$\pm$1.0   & 79.6$\pm$1.2   & 64.4$\pm$1.3 & 70.17 \\
GraphCL      & 69.68$\pm$0.67 & 73.87$\pm$0.66 & 62.40$\pm$0.57 & 60.53$\pm$0.88 & 75.99$\pm$2.65 & 69.80$\pm$2.66 & \textbf{78.47$\pm$1.22} & 75.38$\pm$1.44  & 67.88$\pm$0.85 & 70.44 \\ \hline
SEGA & \textbf{71.86$\pm$1.06} & \textbf{76.72$\pm$0.43} & \textbf{65.23$\pm$0.91} & \textbf{63.68$\pm$0.34} & \textbf{84.99$\pm$0.94} & \textbf{76.60$\pm$2.45} & 77.63$\pm$1.37 & 77.07$\pm$0.46 & \textbf{68.73$\pm$0.54} & \textbf{73.61} \\
\hline \hline
\end{tabular}%
}
\end{small}
\vskip -0.1in
\end{table*}

\section{Experiments}
In this section, we are devoted to evaluating SEGA with extensive experiments~\footnote{The code of SEGA is available at \url{https://github.com/Wu-Junran/SEGA}.}. 
Note that the proposed anchor view is orthogonal to previous works for graph augmentations, and this also reveals that our method has a superior collaborative capability with previous methods.
Therefore, we first validate SEGA via contrasting with the well-known rules for graph augmentations from GraphCL (the first graph contrastive learning method with augmentations)~\cite{you2020graph}.
Then, thorough orthogonal experiments are performed to show the superiority of SEGA against SOTA competitors.
Further ablation studies are conducted to make an in-depth analysis of SEGA.

\subsection{Contrastive Learning with Simple Rules}
\textbf{Datasets.}
For unsupervised and semi-supervised learning, various benchmarks are adopted from TUDataset~\cite{morris2020tudataset}, including COLLAB, REDDIT-BINARY, REDDIT-MULTI-5K, IMDB-BINARY, GITHUB, NCI1, MUTAG, PROTEINS and DD.
For transfer learning, ZINC15 \cite{sterling2015zinc} dataset is adopted for biochemical pre-training. In particular, a subset with two million unlabeled molecular graphs are sampled from the ZINC15.
For protein domain, following \citet{hu2020strategies}, 306K unlabeled protein ego-networks are utilized for pre-training. 
We employ the eight ubiquitous benchmarks from the MoleculeNet dataset \cite{wu2018moleculenet} as the biochemical downstream experiments. 
The protein downstream task is to predict 40 fine-grained biological functions of 8 species.
Further details are shown in Appendix C.

\textbf{Learning protocol.}
Following the learning setting in GraphCL~\cite{you2020graph}, the corresponding learning protocols are adopted for a fair comparison.
(a) In unsupervised representation learning, all data is used for model pre-training and the learned graph embeddings are then fed into a non-linear SVM classifier to perform 10-fold cross-validation.
(b) In transfer learning, we first pre-train the model on ZINC15 and PPI306K. Then, we finetune and evaluate the model on MoleculeNet dataset and PPI using the scaffold split scheme~\cite{chen2012comparison}.
(c) In semi-supervised learning, there exist two learning settings. For datasets with a public training/validation/test split, pre-training is performed only on training dataset, finetuning is conducted with 10\% of the training data, and final evaluation results are from the validation/test sets. 
For datasets without such splits, all samples are employed for pre-training while finetuning and evaluation are performed over 10 folds.

\textbf{Configuration.}
To keep in line with GraphCL~\cite{you2020graph}, the same GNN architectures are employed with their original hyper-parameters under individual experiment settings. 
Specifically, in unsupervised learning, GIN~\cite{xu2019powerful} with 32 hidden units and 3 layers is set up. In addition, the same data augmentations on graphs with the default augmentation strength 0.2 are adopted.
In transfer learning, GIN is used with 5 layers and 300 hidden dimensions.
In semi-supervised learning, ResGCN with 128 hidden units and 5 layers is set up for pre-training and finetuning.

As for the anchor view representation learning, the number of tree encoder layer is consistent with the tree height, which ranges from 2 to 5 and the MLP in each iteration has 2 layers. The encoder hidden dimensions are fixed for all layers to keep in line with GraphCL under individual experiment setting. Additional details are shown in the Appendix D.

\begin{table*}[!t]
\centering
\caption{Average accuracies (\%) $\pm$ Std. of compared methods via semi-supervised representation learning with 10\% labels. \textbf{Bold} indicates the best performance over all methods. A.A. is short for average accuracy. The results of baselines are derived from the published works.}
\label{tab:semisupervised}
\vskip 0.15in
\begin{small}
\resizebox{\textwidth}{!}{%
\begin{tabular}{lccc|cccc|c}
\hline \hline
 & NCI1 & PROTEINS & DD & COLLAB & RED-B & RED-M5K & GITHUB & A.A. \\ \hline \hline
No Pre-Train & 73.72$\pm$0.24 & 70.40$\pm$1.51 & 73.56$\pm$0.41 & 73.71$\pm$0.27 & 86.63$\pm$0.27 & 51.33$\pm$0.44 & 60.87$\pm$0.17 & 70.03 \\
GAE & 74.36$\pm$0.24 & 70.51$\pm$0.17 & 74.54$\pm$0.68 & 75.09$\pm$0.19 & 87.69$\pm$0.40 & 53.58$\pm$0.13 & 63.89$\pm$0.52 & 71.38 \\
ContextPred & 73.00$\pm$0.30 & 70.23$\pm$0.63 & 74.66$\pm$0.51 & 73.69$\pm$0.37 & 84.76$\pm$0.52 & 51.23$\pm$0.84 & 62.35$\pm$0.73 & 69.99 \\
Infomax & 74.86$\pm$0.26 & 72.27$\pm$0.40 & 75.78$\pm$0.34 & 73.76$\pm$0.29 & 88.66$\pm$0.95 & 53.61$\pm$0.31 & 65.21$\pm$0.88 & 72.02 \\
GraphCL & 74.63$\pm$0.25 & 74.17$\pm$0.34 & 76.17$\pm$1.37 & 74.23$\pm$0.21 & 89.11$\pm$0.19 & 52.55$\pm$0.45 & 65.81$\pm$0.79 & 72.38 \\ \hline
SEGA & \textbf{75.09$\pm$0.22} & \textbf{74.65$\pm$0.54} & \textbf{76.33$\pm$0.43} & \textbf{75.18$\pm$0.22} & \textbf{89.40$\pm$0.23} & \textbf{53.73$\pm$0.28} & \textbf{66.01$\pm$0.66} & \textbf{72.92}\\ \hline \hline
\end{tabular}%
}
\end{small}
\vskip -0.1in
\end{table*}

\begin{table*}[!t]
\centering
\caption{Orthogonal experiment results (\%) of SEGA with SOTAs in unsupervised representation learning. \textbf{Bold} indicates the best performance within each specific opeartion. A.A. shows the average accuracy over all datasets. The results of baselines are derived from the published works and - indicates the data missing in the such works.}
\label{tab:cooperation}
\vskip 0.15in
\begin{small}
\resizebox{\textwidth}{!}{
\begin{tabular}{l|cccc|ccccc|c}
\hline \hline
 & NCI1 & PROTEINS & DD & MUTAG & COLLAB & RED-B & RED-M5K & IMDB-B & IMDB-M & A.A. \\ \hline \hline
AD-GCL-FIX & 69.67$\pm$0.51 & 73.59$\pm$0.65 & 74.49$\pm$0.52 & 89.25$\pm$1.45 & 73.32$\pm$0.27 & 85.52$\pm$0.79 & 53.00$\pm$0.82 & 71.57$\pm$1.01 & 49.04$\pm$0.53 & 71.05 \\ 
\rowcolor{gray!30} SEGA-AD-GCL-FIX & \textbf{70.38$\pm$0.76} & \textbf{74.61$\pm$0.81} & \textbf{75.84$\pm$0.64} & \textbf{89.89$\pm$0.69} & \textbf{75.03$\pm$0.36} & \textbf{87.74$\pm$0.39} & \textbf{54.29$\pm$0.54} & \textbf{72.32$\pm$0.49} & \textbf{50.83$\pm$0.34} & \textbf{72.33}($\uparrow$1.28) \\ 
JOAO & \textbf{78.07$\pm$0.47} & 74.55$\pm$0.41 & 77.32$\pm$0.54 & 87.35$\pm$1.02 & 69.50$\pm$0.36 & 85.29$\pm$1.35 & 55.74$\pm$0.63 & 70.21$\pm$3.08 & - & 74.75 \\ 
\rowcolor{gray!30} SEGA-JOAO & 76.19$\pm$0.77 & \textbf{75.44$\pm$0.54} & \textbf{78.27$\pm$1.32} & \textbf{87.70$\pm$1.31} & \textbf{72.82$\pm$0.35} & \textbf{86.79$\pm$1.31} & \textbf{56.17$\pm$0.67} & \textbf{71.74$\pm$1.26} & - & \textbf{75.64}($\uparrow$0.89) \\ 
JOAOv2 & \textbf{78.36$\pm$0.53} & 74.07$\pm$1.10 & 77.40$\pm$1.15 & 87.67$\pm$0.79 & 69.33$\pm$0.34 & 86.42$\pm$1.45 & 56.03$\pm$0.27 & 70.83$\pm$0.25 & - & 75.01 \\ 
\rowcolor{gray!30} SEGA-JOAOv2 & 78.04$\pm$0.19 & \textbf{75.94$\pm$0.88} & \textbf{78.37$\pm$1.26} & \textbf{88.53$\pm$2.45} & \textbf{72.76$\pm$0.27} & \textbf{87.98$\pm$0.29} & \textbf{56.15$\pm$0.29} & \textbf{72.12$\pm$0.79} & - & \textbf{76.24}($\uparrow$1.23) \\ 
AutoGCL & \textbf{82.00$\pm$0.29} & 75.80$\pm$0.36 & 77.57$\pm$0.60 & 88.64$\pm$1.08 & 70.12$\pm$0.68 & 88.58$\pm$1.49 & 56.75$\pm$0.18 & 73.30$\pm$0.40 & - & 76.59 \\ 
\rowcolor{gray!30} SEGA-AutoGCL & 81.84$\pm$0.53 & \textbf{76.43$\pm$0.67} & \textbf{78.31$\pm$1.37} & \textbf{89.03$\pm$1.01} & \textbf{72.68$\pm$0.23} & \textbf{89.88$\pm$1.21} & \textbf{57.43$\pm$0.37} & \textbf{73.95$\pm$0.87} & - & \textbf{77.44}($\uparrow$0.85) \\ 
RGCL & 78.14$\pm$1.08 & 75.03$\pm$0.43 & 78.86$\pm$0.48 & 87.66$\pm$1.01 & 70.92$\pm$0.65 & 90.34$\pm$0.58 & 56.38$\pm$0.40 & 71.85$\pm$0.84 & - & 76.15 \\ 
\rowcolor{gray!30} SEGA-RGCL & \textbf{79.42$\pm$0.82} & \textbf{75.87$\pm$0.45} & \textbf{79.54$\pm$1.14} & \textbf{88.79$\pm$1.87} & \textbf{73.14$\pm$0.37} & \textbf{90.75$\pm$0.84} & \textbf{57.28$\pm$0.42} & \textbf{72.75$\pm$0.66} & - & \textbf{77.19}($\uparrow$1.04) \\ \hline \hline
\end{tabular}
}
\end{small}
\vskip -0.1in
\end{table*}

\textbf{Unsupervised learning.}
The compared methods in unsupervised learning have three categories. The published hyper-parameters of these methods are adopted. The first set is three SOTA kernel-based methods that include GL~\cite{shervashidze2009efficient}, WL~\cite{shervashidze2011weisfeiler}, and DGK~\cite{yanardag2015deep}. The second set is four heuristic self-supervised methods, including node2vec \cite{grover2016node2vec}, sub2vec \cite{adhikari2018sub2vec}, graph2vec \cite{narayanan2017graph2vec}, and InfoGraph \cite{sun2020infograph}. The final compared methods are MVGRL \cite{hassani2020contrastive} and GraphCL \cite{you2020graph} for unsupervised learning.

The classification accuracies of SEGA contrasted with simple augmentation rules under the setting of unsupervised learning are shown in~\cref{tab:unsupervised}, and a significant performance improvement from the appearance of the target anchor view can be witnessed as opposed to the baselines.
Specifically, in light of the last column for average rank, SEGA acquires the highest position among the ten methods.
Moreover, as can be seen from the column for average accuracy, SEGA outperforms InfoGraph and GraphCL with a 3.22\% and 1.54\% accuracy gain.
In particular, except for the performance on NCI1, SEGA obtains the best performance on the other seven benchmarks, and we still can observe that SEGA obtains the highest accuracy over all eight benchmarks under the scenario without kernel-based methods. 
Thus, we can conclude that better performance can consistently be achieved when contrasting with the proposed anchor view.

\textbf{Transfer learning.}
The baseline methods under the setting of transfer learning include EdgePred, AttrMsking, ContexPred \cite{hu2020strategies}, Infomax \cite{velickovic2019deep} and GraphCL \cite{you2020graph}. A model without pre-train, termed `No Pre-Train', is also adopted for comparison.

The results of SEGA, along with baselines under the setting of transfer learning on nine benchmarks, are shown in~\cref{tab:trans-results}. 
To summarize, the proposed anchor view, SEGA, obtains superior performance compared to previous works.
Specifically, SEGA achieves the best performance on seven out of nine benchmarks, and a 3.17\% performance gain is obtained in terms of average ROC-AUC compared to GraphCL.
Thus, we can conclude that the proposed anchor view servers as a good contrastive branch to help the graph encoder model essential information of given graphs and improve generalization and performance.

\textbf{Semi-supervised learning.}
Five baselines are adopted for semi-supervised learning, including a naive GCN without pre-training~\cite{you2020graph}, GAE~\cite{kipf2016variational}, Infomax~\cite{velickovic2019deep}, ContextPred~\cite{hu2020strategies} and GraphCL~\cite{you2020graph}.

The classification accuracies of SEGA and compared methods under the setting of semi-supervised learning are shown in~\cref{tab:semisupervised}, and SEGA outperforms these compared methods across all benchmarks.
Despite the least performance improvement, the effectiveness of our proposed anchor view still has been validated in semi-supervised learning.

\subsection{Orthogonal to SOTAs}
As mentioned above, the proposed anchor view is orthogonal to previous works for graph augmentations; thus, we further evaluate SEGA in collaboration with these augmented views in unsupervised learning setting, including AD-GCL~\cite{suresh2021adversarial}, JOAO~\cite{you2021graph}, AutoGCL~\cite{yin2022autogcl} and RGCL~\cite{li2022let}. Detailed settings for orthogonal experiments are shown in Section~\ref{sec:setting-ortho}. 

The orthogonal results are shown in~\cref{tab:cooperation}, and we can see that a general performance improvement is achieved with the SEGA. Despite several specific failures, 0.85\%$\sim$1.28\% average accuracy gains confirm the effectiveness of SEGA as an anchor view for graph contrastive learning.

\begin{figure}[!ht]
\vskip 0.2in
\begin{center}
\centerline{\includegraphics[width=\linewidth]{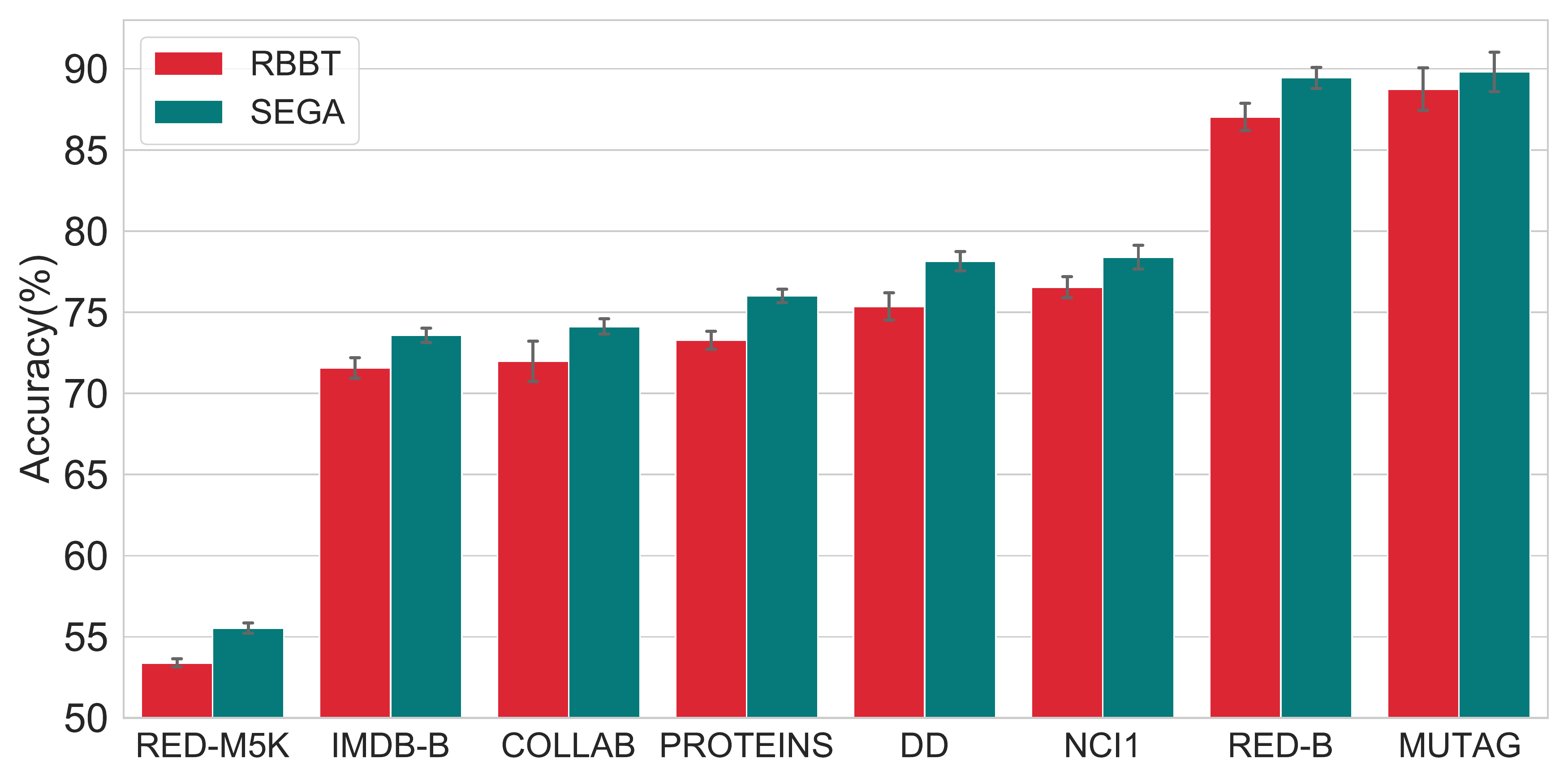}}
\caption{Performance comparison between RBBT and SEGA.}
\label{fig:rbbt-sega}
\end{center}
\vskip -0.2in
\end{figure}

\subsection{Ablation Study}
\label{sec:ablation}
Here, we make an in-depth analysis about the performance of SEGA under the setting of unsupervised learning.

\textbf{Guidance from structural entropy.}
Besides the superior performance of SEGA, we further evaluate the effectiveness of~\cref{code:coding_tree} for structural uncertainty minimization. In unsupervised learning, we produce the anchor view without guidance from structural entropy but adopt a random coding tree, i.e., a randomly balanced binary tree (RBBT) with a height of two.
We also fix the height of the guided anchor view to two for fair comparison.
The results are shown in \cref{fig:rbbt-sega} and the structural entropy-guided anchor view surpasses the random coding tree on all eight benchmarks.

\begin{figure}[!t]
\vskip 0.2in
\begin{center}
\centerline{\includegraphics[width=\linewidth]{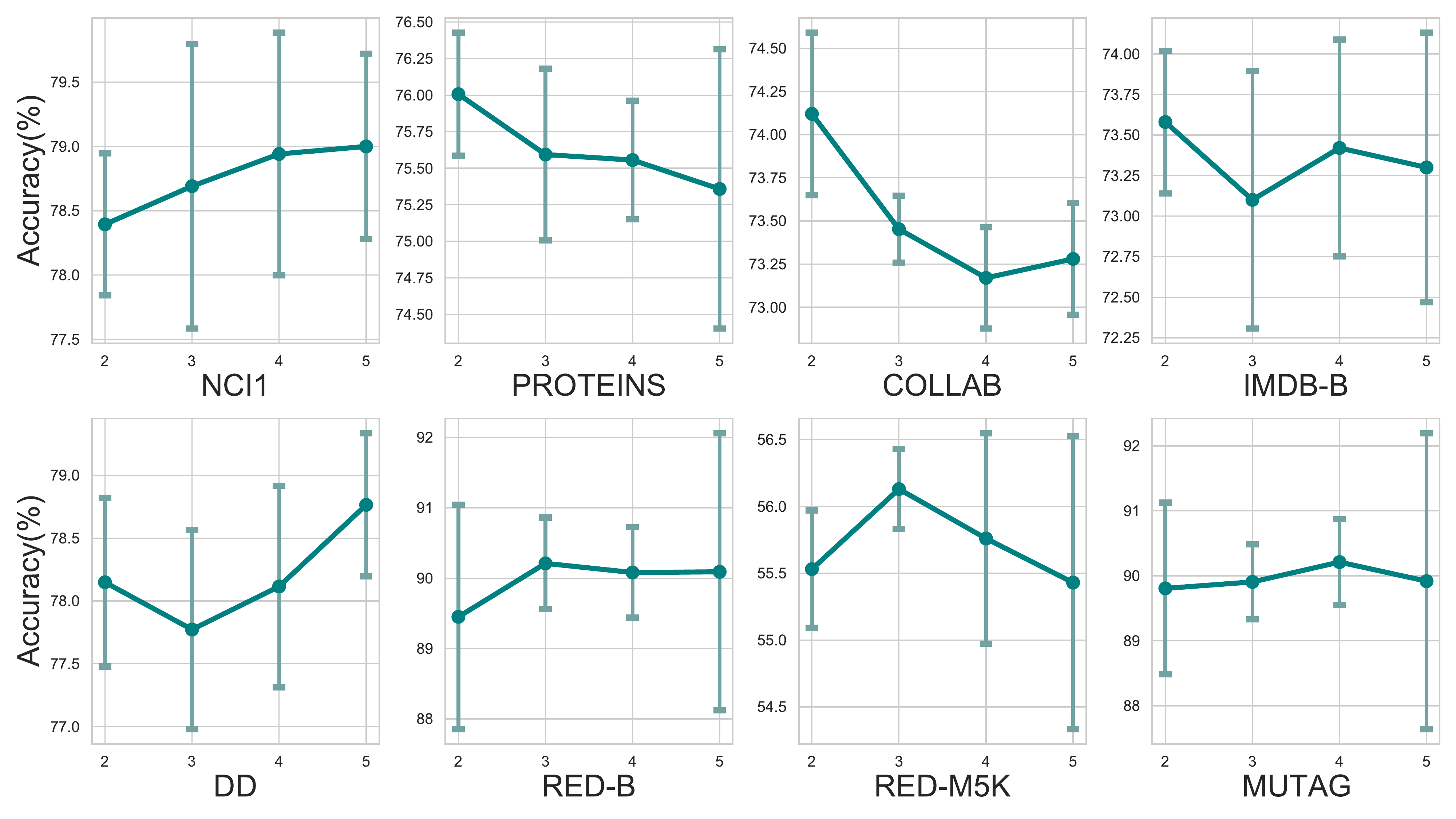}}
\caption{The natural hierarchy of graph.}
\label{fig:sega-height}
\end{center}
\vskip -0.2in
\end{figure}

\textbf{The height $k$ of graph's natural hierarchy.}
In experimental setup, the height $k$ of coding tree ranges from 2 to 5. Here, we delve deeper into the optimal height $k$ of graph's natural hierarchy. The specific performance of SEGA under each height $k$ via unsupervised learning is shown in~\cref{fig:sega-height}. As can be seen, the optimal height $k$ with the highest accuracy varies among datasets. Except for NCI1 and DD, the other six benchmarks achieve the best performance with shallow layers (less than 5), and the rising trend of NCI1 and DD also implies the great potential of SEGA.

\section{Conclusion}
In this work, we try to explore an anchor view that maintains the essential information of input graphs for graph contrastive learning. In the light of the graph information bottleneck, we attempt to give the definition of the expected anchor view.
Moreover, based on the structural information theory, we present a practical instantiation, called SEGA, to implement this anchor view for graph contrastive learning.  
Contrasted with extensive views in previous works, SEGA shows superior performance on tasks regarding graph classification compared to SOTAs. 
An anchor view that maintains the essential information for node classification sheds light on our future research direction.

\section*{Acknowledgements}
This research was supported by NSFC (Grant No. 61932002).

\bibliography{example_paper}
\bibliographystyle{icml2023}

\newpage
\appendix
\onecolumn

\setcounter{table}{0}
\setcounter{figure}{0}
\renewcommand{\thetable}{A.\arabic{table}}
\renewcommand{\thefigure}{A.\arabic{figure}}




\section{Proof of $\mathcal{H}(G^\ast|G)=0$}

\begin{figure}[!ht]
  \centering
  \begin{subfigure}{.4\linewidth}
    \centering
    \includegraphics[width=\linewidth]{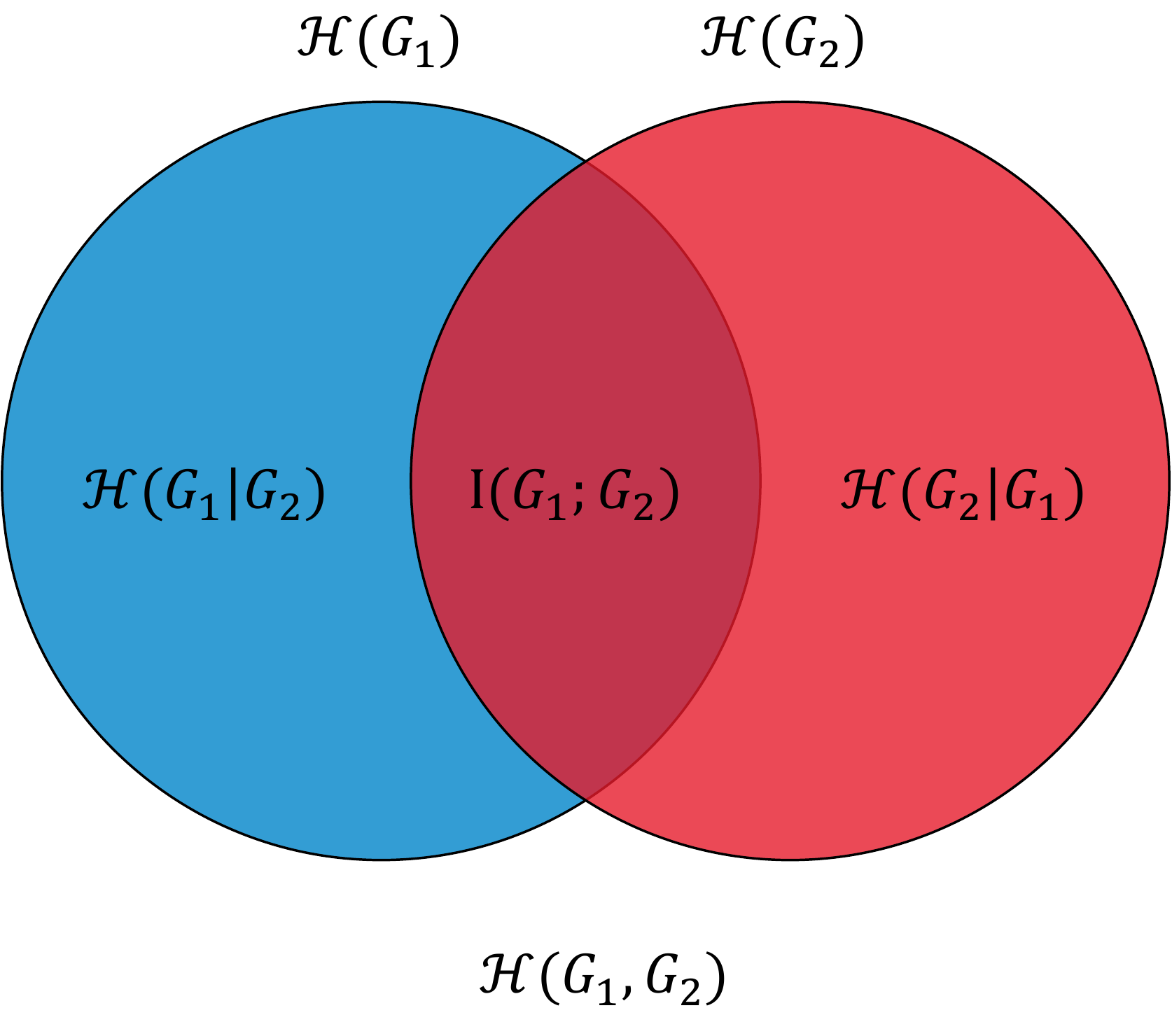}
    \caption{MI with data augmentations.}
    \label{fig:mi-v1v2}
  \end{subfigure} 
  \begin{subfigure}{.4\linewidth}
    \centering
    \includegraphics[width=\linewidth]{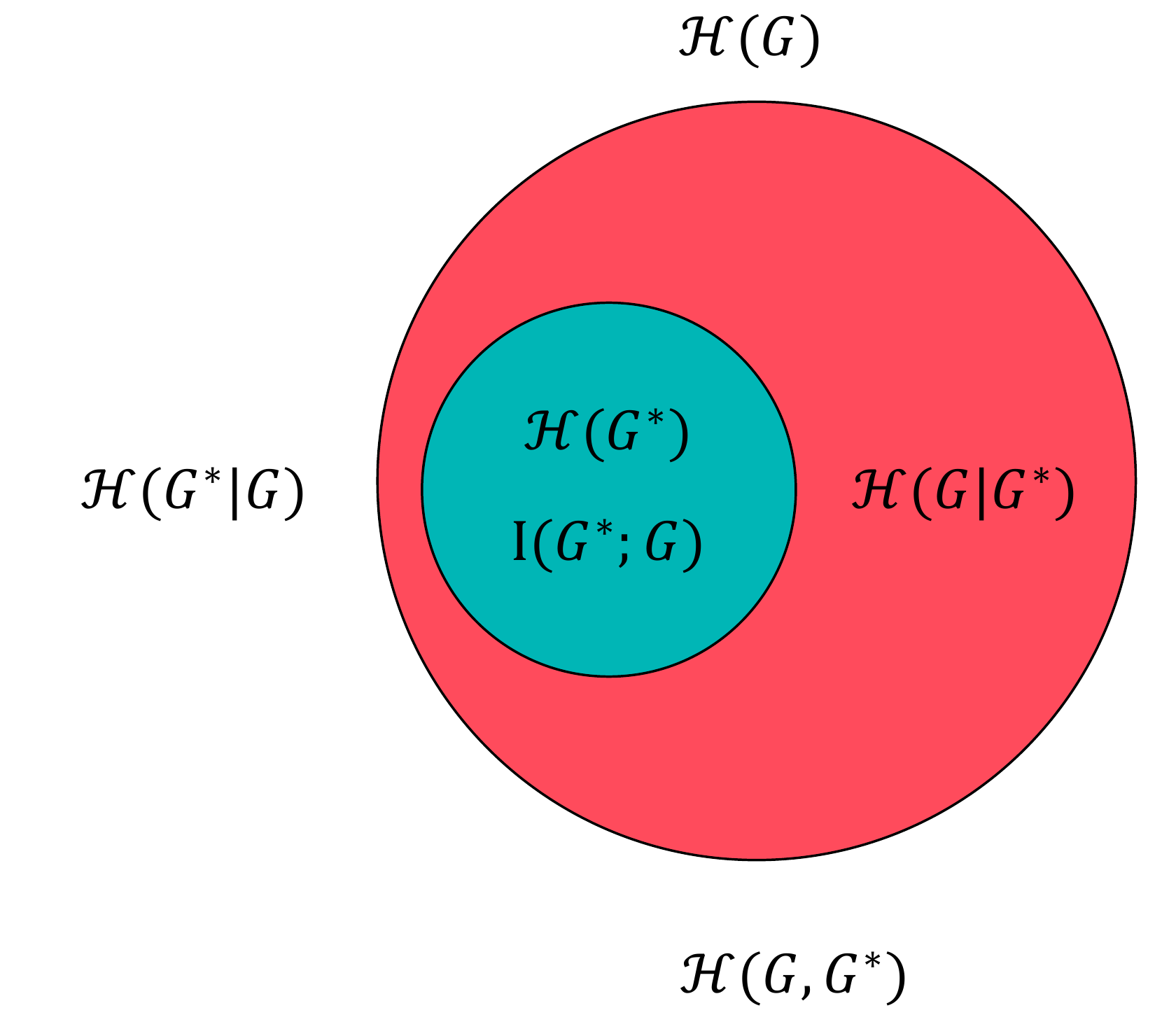}
    \caption{MI with the anchor view.}
    \label{fig:mi-v}
  \end{subfigure} 
  \caption{Mutual information with data augmentations and the target anchor view.}
  \label{fig:mi}
\end{figure}

In this section, we present the proof of the statement $\mathcal{H}(G^\ast|G)=0$. First, we repeat the property that the target anchor view should own:
\begin{definition}
The anchor view with essential information is supposed to be a substructure of the given graph to avoid artificially introduced noise.
\label{def:view_property_app}
\end{definition}
Before the proof, Figure~\ref{fig:mi} first shows the mutual information of graphs with data augmentations and the target anchor view (i.e., comply with Definition~\ref{def:view_property_app}). Figure~\ref{fig:mi-v1v2} suggests that the mutual information with data augmentations is the common part of two views. While Figure~\ref{fig:mi-v} reveals that the mutual information with the target anchor view is the information within the essential part.

\begin{theorem}
Suppose the target anchor view $G^\ast$ of the corresponding graph $G$ owns the property in Definition~\ref{def:view_property_app}, the mutual information between $G^\ast$ and $G$ is
\begin{align}
I(G^\ast; G) =&\,\, \mathcal{H}(G^\ast) - \mathcal{H}(G^\ast|G) \nonumber \\
=&\,\, \mathcal{H}(G^\ast).
\end{align}
\end{theorem}

\begin{proof}
According to the definition of Shannon entropy, i.e., $\mathcal{H}(X) = -\mathop{\sum}\limits _{x\in X} P(x)\log P(x)$, we follow the formulation of graph mutual information in \citet{sun2020infograph} that $X$ is a set of node representations drawn from an empirical probability distribution of graph $G$, so the conditional entropy can be written as
\begin{align}
\label{eq:1}
\mathcal{H}(G^\ast|G) 
  = & \,\,\mathcal{H}(X^\ast|X) \nonumber \\ 
  = & \mathop{\sum}\limits_{x\in X} P(x)H(X^\ast | X=x) \nonumber \\
  = & -\mathop{\sum}\limits_{x\in X} P(x) \mathop{\sum}\limits_{x^\ast \in X^\ast} P(x^\ast|x)\log P(x^\ast|x) \nonumber \\
  = & -\mathop{\sum}\limits_{x\in X} \mathop{\sum}\limits_{x^\ast \in X^\ast} P(x^\ast, x)\log P(x^\ast|x) \nonumber \\
  = & -\mathop{\sum}\limits_{x^\ast, x} P(x^\ast, x)\log P(x^\ast | x).
\end{align}
Considering that $G^\ast$ complies with the Definition~\ref{fig:mi-v}, the illustration of probability distribution of $G^\ast$ and $G$ is shown Figure~\ref{fig:mi-v}. Here, let us firstly discuss that
\begin{itemize}

\item when $x \in X$ and $x \notin X^\ast$, we have $P(x^\ast, x)=0$.

\end{itemize}

Therefore, we can transform Equation \ref{eq:1} to

\begin{align}
\label{eq:2}
\mathcal{H}(X^\ast|X)
  =\,\, & -\mathop{\sum}\limits_{x^\ast, x} P(x^\ast, x)\log P(x^\ast | x) \nonumber \\
  =\,\, & -\mathop{\sum}\limits_{x^\ast, x^\ast} P(x^\ast, x^\ast)\log P(x^\ast | x^\ast) \nonumber\\
        & -\mathop{\sum}\limits_{x^\ast, x \notin X^\ast} P(x^\ast, x)\log P(x^\ast | x)  \nonumber \\
  =\,\, & -\mathop{\sum}\limits_{x^\ast \in X^\ast} \mathop{\sum}\limits_{x^\ast \in X^\ast} P(x^\ast, x^\ast)\log P(x^\ast|x^\ast) \nonumber \\
  =\,\, & -\mathop{\sum}\limits_{x^\ast \in X^\ast} P(x^\ast) \mathop{\sum}\limits_{x^\ast \in X^\ast} P(x^\ast|x^\ast)\log P(x^\ast|x^\ast) \nonumber \\ 
  =\,\, & \mathop{\sum}\limits_{x^\ast \in X^\ast} P(x^\ast)H(X^\ast | X^\ast=x^\ast) \nonumber \\
  =\,\, & \mathcal{H}(X^\ast|X^\ast) \nonumber \\
  =\,\, & 0.
\end{align}

Therefore, given $\forall x \in X$, we have $\mathcal{H}(G^\ast | G)= \mathcal{H}(X^\ast|X) = 0$. Accordingly, we have
\begin{equation}
I(G^\ast; G) = \,\, \mathcal{H}(G^\ast).
\end{equation}
\end{proof}

\section{Illustrations for Algorithm 1}
\begin{figure}[!ht]
  \centering
  \begin{subfigure}{0.48\textwidth}
    \centering
    \includegraphics[width=.6\linewidth]{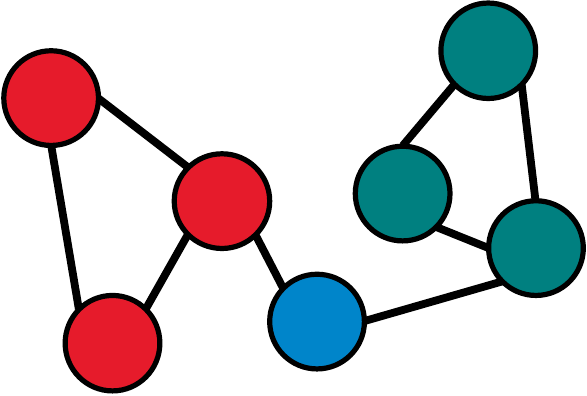}
  \end{subfigure}
  \begin{subfigure}{.48\textwidth}
    \centering
    \includegraphics[width=\linewidth]{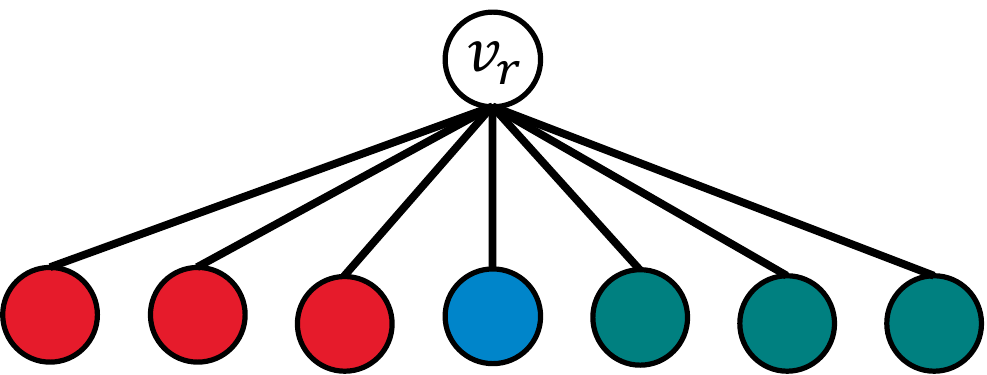}
  \end{subfigure} 
  \\
  \begin{subfigure}{.48\textwidth}
    \centering
    \caption{Original Graph $G$.}
  \end{subfigure}
  \begin{subfigure}{.48\textwidth}
    \centering
    \caption{Initial Coding Tree of $G$.}
  \end{subfigure} 
  \caption{The original graph $G$ and the initialized coding tree with only root node and vertices from $G$ as leaves.} 
  \label{fig:preparation}
\end{figure}

Here, we present several figures to clearly reveal the running process of function COMBINE$(\cdot)$ and DROP$(\cdot)$ needed by Algorithm 1. Moreover, the growing process of a coding tree with a fixed height 2 from its original graph is further presented in Figure~\ref{fig:algorithm1}. First, as shown in Figure~\ref{fig:preparation}, we give a simple undirected graph $G=\{\mathcal{V}, \mathcal{E}\}$ for structural entropy minimization and the corresponding initialized coding tree $T$ from $G$. Here, we present the definition of coding tree.

\paragraph{Coding Tree.} A coding tree of a simple undirected graph $G=\{\mathcal{V}, \mathcal{E}\}$ is defined as a rooted tree $T$ that has the following properties:
\begin{itemize}
  \item The root node $v_r$ is associated with the vertices set $\mathcal{V}$ of $G$. $v_r$ is termed as the codeword of $\mathcal{V}$, that is $c(V)=v_r$. $\mathcal{V}$ is termed as the marker of $v_r$, that is $M(v_r) = \mathcal{V}$.
  \item Every node $v_\tau\in T$ is a codeword of a subset $\tilde{\mathcal{V}}\subset\mathcal{V}$; put differently, $c(\tilde{\mathcal{V}}) = v_\tau$ and $M(v_\tau) = \tilde{\mathcal{V}}$.
  \item For every node $v_\tau\in T$, suppose that $v^+_1, v^+_2, \cdots, v^+_N$ are all the immediate successors of $v_\tau$ in T; then all $M(v^+_i)$ are disjointed, and $M(v_\tau)=\bigcup_{i=1}^{N}M(v^+_i)$.
  \item For every leaf node $v_\tau^l \in T$, $M(v_\tau^l)$ is a singleton $\{v\}$ for some vertex $v\in\mathcal{V}$, and for every vertex $v\in\mathcal{V}$, there is a unique leaf node $v_\tau^l\in T$ such that $M(v_\tau^l)={v}$ and $c(v)=v_\tau^l$.
\end{itemize}

\subsection{Illustration of COMBINE}
\begin{figure}[!ht]
\centering
\centerline{\includegraphics[width=1.\linewidth]{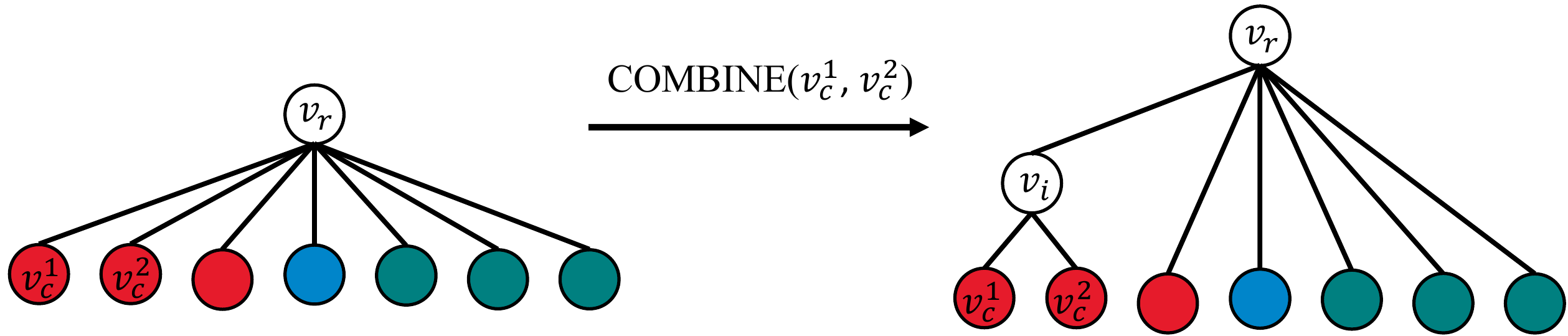}}
\caption{An illustration of COMBINE$(\cdot)$.} 
\label{fig:combine}
\end{figure}

The process of COMBINE$(\cdot)$ is illustrated in Figure~\ref{fig:combine}. Specifically, let $v_c^1$ and $v_c^1$ be any two child nodes of root node $v_r$, then, a virtual node $v_i$ is inserted between the root node and the two children, in which $v_c^1$ and $v_c^1$ become the children of $v_i$ and $v_i$ directly dissolves into the children cluster of $v_r$.

\subsection{Illustration of DROP}
\begin{figure}[!ht]
\centering
\centerline{\includegraphics[width=1.\linewidth]{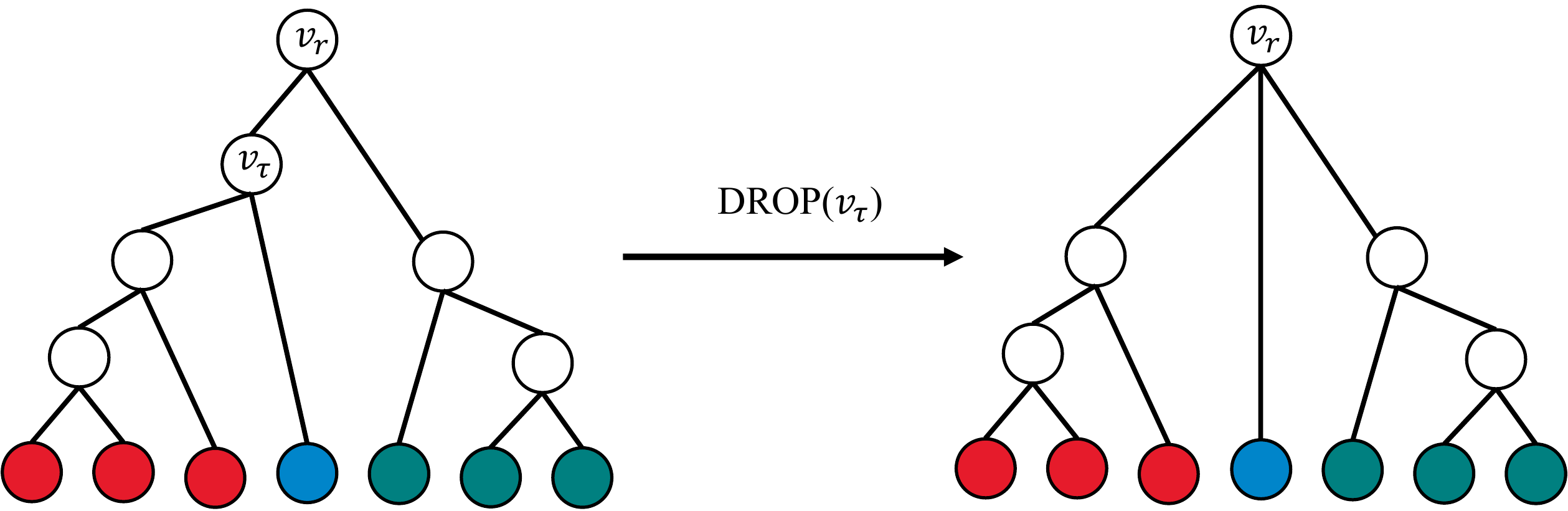}}
\caption{An illustration of DROP$(\cdot)$.} 
\label{fig:drop}
\end{figure}

The process of DROP$(\cdot)$ is illustrated in Figure~\ref{fig:drop}. Specifically, given an inner node $v_\tau$ of the coding tree, then, $v_\tau$ is removed from the tree and its children are adopted by its parent node.

\subsection{Illustration of Algorithm 1}
\begin{figure}[!ht]
  \centering
  \begin{subfigure}{\textwidth}
    \centering
    \includegraphics[width=\linewidth]{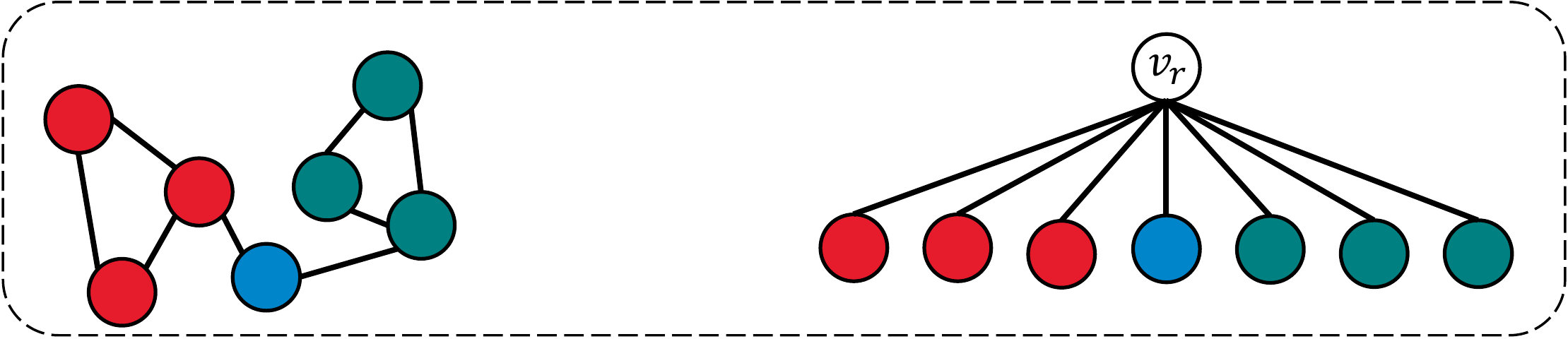}
    \caption{Input and Line 1 of Algorithm 1.}
    \label{fig:al1-st0}
  \end{subfigure}
  \\
  \begin{subfigure}{\textwidth}
    \centering
    \includegraphics[width=\linewidth]{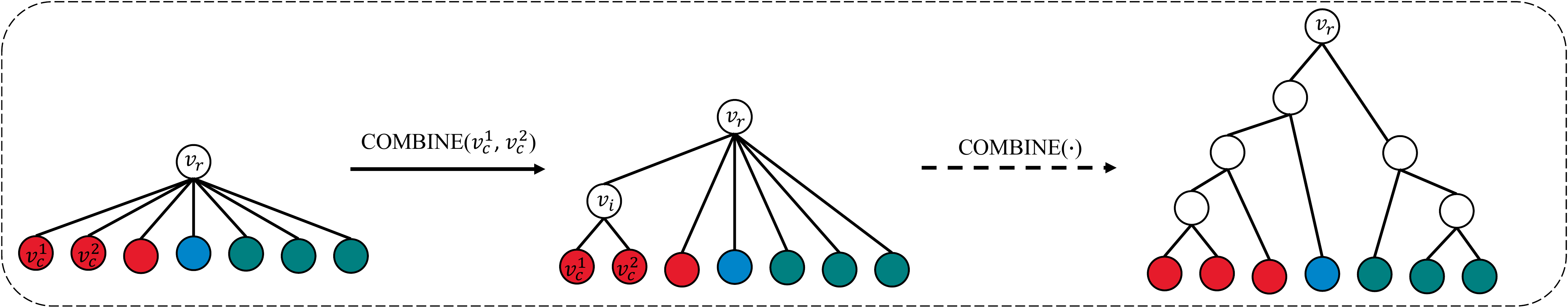}
    \caption{Stage 1 (Line 2-5): Construct a full-height binary coding tree from bottom to top.}
    \label{fig:al1-st1}
  \end{subfigure}
  \\
  \begin{subfigure}{\textwidth}
    \centering
    \includegraphics[width=\linewidth]{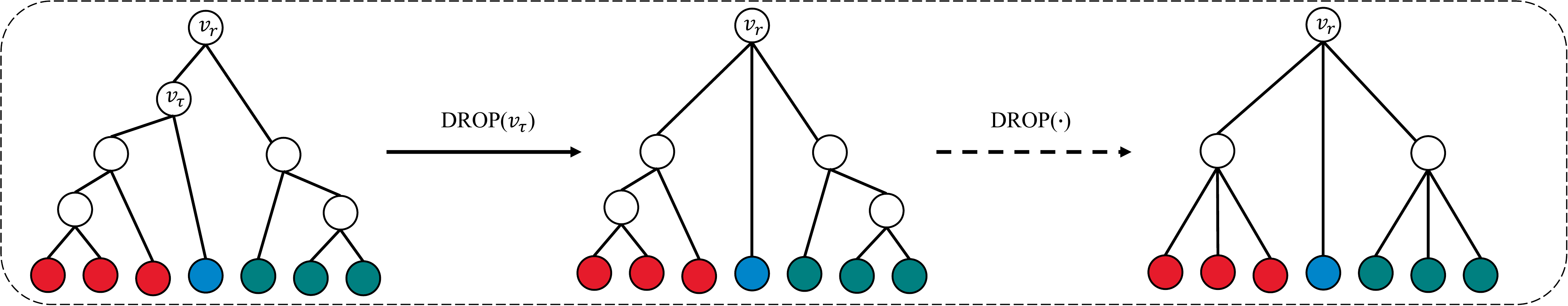}
    \caption{Stage 2 (Line 6-9): Squeeze full-height binary coding tree to fixed height 2.}
    \label{fig:al1-st2}
  \end{subfigure}
  \caption{An illustration of the running process of Algorithm~\ref{code:coding_tree}.} 
  \label{fig:algorithm1}
\end{figure}

The running process of Algorithm~\ref{code:coding_tree} is illustrated in Figure~\ref{fig:algorithm1} and we set the target height of coding tree to 2. The input graph and the initialized coding tree are shown in Figure~\ref{fig:al1-st0}, in which the coding tree is initialized with a root node $v_r$ and all vertices from input graph as leaves. Figure~\ref{fig:al1-st1} shows the process of Stage 1. Through iteratively combining two children of the root node, which can achieve the maximal structural entropy reduction after combination, a full-height binary coding tree is weaved from bottom to top. Figure~\ref{fig:al1-st2} reveals the process of State 2. Each time, an inner node that achieves the minimal structural entropy restoration is dropped. Finally, a coding tree with height of 2 can be harvested as a view for graph contrastive learning.

\paragraph{Complexity analysis.} Given a graph $G=(V, E)$, $|V|=n$ and $|E|=m$, the runtime complexity of Algorithm 1 is $O(h_{max}(m\log n+n))$, in which $h_{max}$ is the height of coding tree $T$ after the first stage. In general, the coding tree $T$ tends to be balanced in the process of structural entropy minimization, thus, $h_{max}$ will be around $\log n$. Furthermore, a graph generally has more edges than nodes, i.e., $m\gg n$, thus the runtime of Algorithm 1 almost scales linearly in the number of edges.

\section{Summary of Datasets}

\subsection{Datasets for Unsupervised and Semi-supervised Learning}
A wide variety of datasets from different domains for a range of graph property prediction tasks are used for our experiments. Here, we present detailed descriptions of the 10 benchmarks utilized in this paper. Table~\ref{tab:data_stat} shows statistics for datasets. 

\paragraph{Social Network Datasets.} IMDB-BINARY and IMDB-MULTI are derived from the collaboration of a movie set. In these two datasets, every graph consists of actors or actresses, and each edge between two nodes represents their cooperation in a certain movie. Each graph is derived from a prespecified movie, and its label corresponds to the genre of this movie. Similarly, COLLAB is also a collaboration dataset but from a scientific realm, which includes three public collaboration datasets (i.e., Astro Physics, High Energy Physics and Condensed Matter Physics). Many researchers from each field form various ego networks for the graphs in this benchmark. The label of each graph is the research field to which the nodes belong. REDDIT-BINARY and REDDIT-MULTI-5K are balanced datasets, where each graph corresponds to an online discussion thread and nodes correspond to users. An edge is drawn between two nodes if at least one of them responds to another's comment. The task is to classify each graph into the community or subreddit to which it belongs.

\paragraph{Small Molecules.} NCI1 is a dataset made publicly available by the National Cancer Institute (NCI) and is a subset of balanced datasets containing chemical compounds screened for their ability to suppress or inhibit the growth of a panel of human tumor cell lines; this dataset possesses 37 discrete labels. 
MUTAG has seven kinds of graphs that are derived from 188 mutagenic aromatic and heteroaromatic nitro compounds. PTC includes 19 discrete labels and reports the carcinogenicity of 344 chemical compounds for male and female rats.

\paragraph{Bioinformatic Datasets.} DD contains graphs of protein structures. A node represents an amino acid and edges are constructed if the distance of two nodes is less than $6\AA$. A label denotes whether a protein is an enzyme or non-enzyme. 
PROTEINS is a dataset where the nodes are secondary structure elements (SSEs), and there is an edge between two nodes if they are neighbors in the given amino acid sequence or in 3D space. The dataset has 3 discrete labels, representing helixes, sheets or turns.

\begin{table}[!ht]
\centering
\caption{Statistics for datasets of diverse nature from the benchmark TUDataset.}
\label{tab:data_stat}
\begin{tabular}{l|cccc}
\hline \hline
Dataset & \#Graphs & \#Classes & Avg. \#Nodes & Avg. \#Edges \\ \hline \hline
\multicolumn{5}{c}{Social Networks} \\ \hline
COLLAB & 5,000 & 3 & 74.49 & 2457.78 \\
REDDIT-BINARY & 2,000 & 2 & 429.63 & 497.75 \\
REDDIT-MULTI-5K & 4,999 & 5 & 508.52 & 594.87 \\
IMDB-BINARY & 1,000 & 2 & 19.77 & 96.53 \\
IMDB-MULTI & 1,500 & 3 & 13.00 & 65.94 \\
GITHUB & 12,725 & 2 & 113.79 & 234.64 \\ \hline
\multicolumn{5}{c}{Small Molecules} \\ \hline
NCI1 & 4,110 & 2 & 29.87 & 32.30 \\
MUTAG & 188 & 2 & 17.93 & 19.79 \\ \hline
\multicolumn{5}{c}{Bioinformatics} \\ \hline
PROTEINS & 1,113 & 2 & 39.06 & 72.82 \\
DD & 1,178 & 2 & 284.32 & 715.66 \\ \hline \hline
\end{tabular}%
\end{table}

\subsection{Details of Molecular Datasets}
\paragraph{Input graph representation.} For simplicity, we use a minimal set of node and bond features that unambiguously describe the two-dimensional structure of molecules. We use RDKit \cite{landrum2013rdkit} to obtain these features.

\begin{itemize}
    \item Node features:
    \begin{itemize}
        \item Atom number: [1, 118]
        \item Chirality tag: \{unspecified, tetrahedral cw, tetrahedral ccw, other\}
    \end{itemize}
    \item Edge features:
    \begin{itemize}
        \item Bond type: \{single, double, triple, aromatic\}
        \item Bond direction: \{–, endupright, enddownright\}
    \end{itemize}
\end{itemize}

\paragraph{Downstream task datasets.} 8 binary graph classification datasets from MoleculeNet \cite{wu2018moleculenet} are used to evaluate model performance.

\begin{itemize}
    \item BBBP \cite{martins2012bayesian}. Blood-brain barrier penetration (membrane permeability), involves records of whether a compound carries the permeability property of penetrating the blood-brain barrier.
    \item Tox21 \cite{Tox21}. Toxicity data on 12 biological targets, which has been used in the 2014 Tox21 Data Challenge and includes nuclear receptors and stress response pathways.
    \item ToxCast \cite{richard2016toxcast}. Toxicology measurements based on over 600 in vitro high-throughput screenings.
    \item SIDER \cite{kuhn2016sider}. Database of marketed drugs and adverse drug reactions (ADR), grouped into 27 system organ classes and also known as the Side Effect Resource.
    \item ClinTox \cite{novick2013sweetlead,gayvert2016data}. Qualitative data classifying drugs approved by the FDA and those that have failed clinical trials for toxicity reasons.
    \item MUV \cite{gardiner2011effectiveness}. Subset of PubChem BioAssay by applying a refined nearest neighbor analysis, designed for validation of virtual screening techniques.
    \item HIV \cite{HIV}. Experimentally measured abilities to inhibit HIV replication.
    \item BACE \cite{subramanian2016computational}. Qualitative binding results for a set of inhibitors of human $\beta$-secretase 1.
\end{itemize}

\paragraph{Details of Dataset Splitting}
For molecular prediction tasks, following \citet{ramsundar2019deep}, we cluster molecules by scaffold (molecular graph substructure) \cite{bemis1996properties}, and recombine the clusters by placing the most common scaffolds in the training set, producing validation and test sets that contain structurally different molecules. Prior work has shown that this scaffold split provides a more realistic estimate of model performance in prospective evaluation compared to random split \cite{chen2012comparison,sheridan2013time}. The split for train/validation/test sets is 80\%:10\%:10\%.

\begin{table}[!t]
\centering
\caption{Datasets statistics summary.}
\label{tab:data-stat-molecule}
\resizebox{\textwidth}{!}{%
\begin{tabular}{l|c|c|cccc}
\hline \hline
Dataset & Category & Utilization & \#Tasks & \#Graphs  & Avg.Node & Avg.Degree \\ \hline \hline
ZINC15  & Biochemical Molecules & Pre-Training & & 2,000,000 & 26.63    & 57.72      \\ 
PPI-306K  & Protein-Protein Intersection Networks & Pre-Training & & 306,925 & 39.82 & 729.62 \\ \hline
BBBP    & Biochemical Molecules & Finetuning & 1       & 2,039     & 24.06    & 51.90      \\
Tox21   & Biochemical Molecules & Finetuning & 12      & 7,831     & 18.57    & 38.58      \\
ToxCast & Biochemical Molecules & Finetuning & 617     & 8,576     & 18.78    & 38.52      \\
SIDER   & Biochemical Molecules & Finetuning & 27      & 1,427     & 33.64    & 70.71      \\
ClinTox & Biochemical Molecules & Finetuning & 2       & 1,477     & 26.15    & 55.76      \\
MUV     & Biochemical Molecules & Finetuning & 17      & 93,087    & 24.23    & 52.55      \\
HIV     & Biochemical Molecules & Finetuning & 1       & 41,127    & 25.51    & 54.93      \\
BACE    & Biochemical Molecules & Finetuning & 1       & 1,513     & 34.08    & 73.71      \\ 
PPI    & Protein-Protein Intersection Networks & Finetuning & 40 & 88,000 & 49.35 &890.77 \\ \hline \hline
\end{tabular}%
}
\end{table}

\subsection{Details of Protein Datasets}
\label{sec:det-pro}
\paragraph{Input graph representation.} The protein subgraphs only have edge features.
\begin{itemize}
    \item Edge features (These edge features indicate whether a particular type of relationship exists between a pair of proteins):
    \begin{itemize}
        \item Neighborhood: \{True, False\}, if a pair of genes are consistently observed in each other’s genome neighborhood
        \item Fusion: \{True, False\}, if a pair of proteins have their respective orthologs fused into a single protein-coding gene in another organism
        \item Co-occurrence: \{True, False\}, if a pair of proteins tend to be observed either as present or absent in the same subset of organisms
        \item Co-expression: \{True, False\}, if a pair of proteins share similar expression patterns
        \item Experiment: \{True, False\}, if a pair of proteins are experimentally observed to physically interact with each other
        \item Database: \{True, False\}, if a pair of proteins belong to the same pathway, based on assessments by a human curator
        \item Text: \{True, False\}, if a pair of proteins are mentioned together in PubMed abstracts
    \end{itemize}
\end{itemize}

\paragraph{Datasets.} A dataset containing protein subgraphs from 50 species is used \cite{zitnik2019evolution}. The original PPI networks do not have node attributes, but contain edge attributes that correspond to the degree of confidence for 7 different types of protein-protein relationships. The edge weights range from 0, which indicates no evidence for the specific relationship, to 1000, which indicates the highest confidence. The weighted edges of the PPI networks are thresholded such that the distribution of edge types across the 50 PPI networks are uniform. Then, for every node in the PPI networks, subgraphs centered on each node were generated by: (1) performing a breadth first search to select the subgraph nodes, with a search depth limit of 2 and a maximum number of 10 neighbors randomly expanded per node, (2) including the selected subgraph nodes and all the edges between those nodes to form the resulting subgraph.

The entire dataset contains 394,925 protein subgraphs derived from 50 species. Out of these 50 species, 8 species (arabidopsis, celegans, ecoli, fly, human, mouse, yeast, zebrafish) have proteins with GO protein annotations. The dataset contains 88,000 protein subgraphs from these 8 species, of which 57,448 proteins have at least one positive coarse-grained GO protein annotation and 22,876 proteins have at least one positive fine-grained GO protein annotation. For the self-supervised pre-training dataset, we use a subset 306,925 protein subgraphs.

Fine-grained protein functions is defined as Gene Ontology (GO) annotations that are leaves in the GO hierarchy, and coarse-grained protein functions is defined as GO annotations that are the immediate parents of leaves \cite{ashburner2000gene,gene2019gene}. For example, a fine-grained protein function is “Factor XII activation”, while a coarse-grained function is ``positive regulation of protein''. The former is a specific type of the latter, and is much harder to derive experimentally. The GO hierarchy information is obtained using GOATOOLS \cite{klopfenstein2018goatools}. The supervised pre-training dataset and the downstream evaluation dataset are derived from the 8 labeled species. The 40-th most common fine-grained protein label only has 121 positively annotated proteins, while the 40-th most common coarse-grained protein label has 9386 positively annotated proteins. This illustrates the extreme label scarcity of the downstream tasks.

\paragraph{Dataset splitting.} In the PPI network, species split simulates a scenario where we have only high-level coarse-grained knowledge on a subset of proteins (prior set) in a species of interest (human in our experiments), and want to predict fine-grained biological functions for the rest of the proteins in that species (test set). For species split, we use 50\% of the protein subgraphs from human as test set, and 50\% as a prior set containing only coarse-grained protein annotations. The protein subgraphs from 7 other labelled species (arabidopsis, celegans, ecoli, fly, mouse, yeast, zebrafish) are used as train and validation sets, which are split 85\% : 15\%. The effective split ratio for the train/validation/prior/test sets is 69\% : 12\% : 9.5\% : 9.5\%.

\section{Detailed Experiment Setup}

\subsection{Settings for Unsupervised Learning}
Following the learning setting in SOTA works, the corresponding learning protocols are adopted for a fair comparison. 
In unsupervised representation learning \cite{sun2020infograph}, all data is used for model pre-training and the learned graph embeddings are then fed into a non-linear SVM classifier to perform 10-fold cross-validation.
Experiments are performed for 5 times each of which corresponds to a 10-fold evaluation as~\cite{sun2020infograph}, with mean and standard deviation of accuracies (\%) reported. As for graph representation learning, models are trained 20 epochs and tested every 10 epochs. Hidden dimension is chosen from $\{32,64\}$, and batch size is chosen from $\{32,128\}$. An Adam optimizer \cite{kingma2015adam} is employed to minimize the contrastive lose with $\{0.01, 0.005, 0.001\}$ learning rate.

\textbf{Data Augmentations on Graphs.}
Follow the data augmentations in GraphCL~\cite{you2020graph}, there are four types of general data augmentations for graph-structured data:
\begin{itemize}
  \item \textbf{Node dropping.} Given the graph $G$, node dropping will randomly discard certain portion of vertices along with their connections. The underlying prior enforced by it is that missing part of vertices does not affect the semantic meaning of $G$. Each node’s dropping probability follows a default i.i.d. uniform distribution (or any other distribution).
  
  \item \textbf{Edge perturbation.} It will perturb the connectivities in $G$ through randomly dropping certain ratio of edges. It implies that the semantic meaning of $G$ has certain robustness to the edge connectivity pattern variances. We also follow an i.i.d. uniform distribution to drop each edge.

  \item \textbf{Attribute masking.} Attribute masking prompts models to recover masked vertex attributes using their context information, i.e., the remaining attributes. The underlying assumption is that missing partial vertex attributes does not affect the model predictions much.

  \item \textbf{Subgraph.} This one samples a subgraph from $G$ using random walk. It assumes that the semantics of $G$ can be much preserved in its (partial) local structure.
\end{itemize}

\subsection{Setting for Semi-supervised Learning}
\paragraph{Configuration.}
ResGCN with 128 hidden units and 5 layers is set up in semi-supervised learning. In addition, the same data augmentations on graphs with the default augmentation strength 0.2 are adopted.
For all datasets we perform experiments with 10\% label rate for 5 times, each of which corresponds to a 10-fold evaluation as~\cite{you2020graph}, with mean and standard deviation of accuracies (\%) reported. For pre-training, learning rate is tuned in $\{0.01, 0.001, 0.0001\}$ and epoch number in $\{20, 40, 60, 80, 100\}$ where grid search is performed. 
For fine-tuning, we following the default setting in \cite{you2020graph}, that is, learning rate is 0.001, hidden dimension is 128, bath size is 128, and the pre-trained models are trained 100 epochs.

\paragraph{Learning protocols.} 
Following the learning setting in SOTA works, the corresponding learning protocols are adopted for a fair comparison. 
In semi-supervised learning \cite{you2020graph}, there exist two learning settings. For datasets with a public training/validation/test split, pre-training is performed only on training dataset, finetuning is conducted with 10\% of the training data, and final evaluation results are from the validation/test sets. 
For datasets without such splits, all samples are employed for pre-training while finetuning and evaluation are performed over 10 folds.

\subsection{Setting for Transfer Learning}

\paragraph{Pre-training dataset.}
ZINC15 \cite{sterling2015zinc} dataset is adopted for biochemical pre-training. In particular, a subset with two million unlabeled molecular graphs are sampled from the ZINC15.
For protein domain, following \citet{hu2020strategies}, 306K unlabeled protein ego-networks are utilized for pre-training.

\paragraph{Pre-training details.} 
In the graph encoder setting in \citet{hu2020strategies}, GIN \cite{xu2019powerful} with five convolutional layers is adopted for message passing. In particular, the hidden dimension is fixed to 300 across all layers and a pooling readout function that averages graph nodes is hired for NT-Xent loss calculation with the scale parameter $\tau = 0.1$. The hidden representations at the last layer are injected into the average pooling function.
An Adam optimizer \cite{kingma2015adam} is employed to minimize the integrated losses produced by the 5-layer GIN encoder. The batch size is set as 256, and all training processes will run 100 epochs.

\paragraph{Fine-tuning dataset.} 
We employ the eight ubiquitous benchmarks from the MoleculeNet dataset \cite{wu2018moleculenet} as the biochemical downstream experiments. These benchmarks include a variety of molecular tasks like physical chemistry, quantum mechanics, physiology, and biophysics. 
The protein downstream task is to predict 40 fine-grained biological functions of 8 species.
For dataset split, the scaffold split scheme \cite{chen2012comparison} is adopted for train/validation/test set generation. 

\paragraph{Fine-tuning details.} 
For downstream tasks, a linear layer is stacked after the pre-trained graph encoders for final property prediction. The downstream model still employs the Adam optimizer for 100 epochs fine-tuning. All experiments on each dataset are performed for ten runs with different seeds, and the results are the averaged ROC-AUC scores (\%) $\pm$ standard deviations. 
The learning rate is selected from \{0.01, 0.001, 0.0001\} and is symmetric for both the encoder and augmenter during self-supervision on the pre-train dataset. To be in line with \cite{you2020graph}, the number of training epochs for pre-training is chosen among \{20, 40, 60, 80, 100\} based on the validation performance on the fine-tune datasets.

\subsection{Settings for Orthogonal Experiment}
\label{sec:setting-ortho}
\noindent \textbf{AD-GCL.} In cooperation with AD-GCL~\cite{suresh2021adversarial}, we faithfully follow the original setting while switching the anchor view from the original graph to the proposed anchor view. Note that, in the evaluation stage, the linear SVM is adopted to keep in line with the results in the main text of AD-GCL, which is different from the setting of GraphCL. In particular, the key hyperparameter $\lambda$ that prevents AD-GCL from very aggressive perturbation is fixed to 5, that is the AD-GCL-FIX in the original work.

\noindent \textbf{JOAO(v2).} In cooperation with JOAO(v2)~\cite{you2021graph}, the same experimental setting is adopted from the published paper while recalling one of the two views and assigning the proposed anchor view to that place. Naturally, JOAO(v2) only needs to search the other view from data augmentations.
Similarly, GIN is adopted for graph encoding while non-linear SVM is employed for evaluation. The hyperparameter $\gamma$ controlling the trade-off between the contrastive loss and view distance is tuned in the range of $\{0.01, 0.1, 1\}$. In particular, JOAO is pre-trained with 20 epochs, while JOAOv2 is pre-trained with double epochs since multiple projection heads are applied.

\noindent \textbf{AutoGCL.} We adopt the naive training strategy proposed in AutoGCL to make a fair comparison.
Specifically, we retain one of the two graph generators and assign our proposed anchor view to the blank position. In particular, AutoGCL extends the layer number of graph encoder from 3 to 5 and the hidden size from 32 to 128. Moreover, AutoGCL is pre-trained with 30 epochs rather than 20 epochs.

\noindent \textbf{RGCL.} In cooperation with RGCL~\cite{li2022let}, we faithfully follow the experiment settings revealed in their codes while replacing one of the two rationale-augmented views with SEGA. Note that, the tuned hyper-parameters in RGCL includes learning rate, sampling ratio $\rho$, loss temperature $\tau$, and loss balance $\lambda$. In particular, RGCL is pre-trained 40 epochs in total and evaluated every 5 epochs.

\end{document}